\documentclass[12pt]{article}

\input{glyphtounicode}  
\usepackage{epstopdf}
\usepackage{blindtext}
\usepackage{graphicx}
\usepackage[tikz]{bclogo}
\usepackage{amsmath}

\usepackage{xspace}	
\usepackage{graphicx}
\usepackage{color}
\definecolor{orange}{RGB}{255,127,0}
\definecolor{brown}{RGB}{150,70,0}
\definecolor{red}{RGB}{255,90,90}
\definecolor{darkred}{RGB}{150,0,0}
\definecolor{myred}{RGB}{200,50,50}
\definecolor{green}{RGB}{127,255,127}
\definecolor{darkgreen}{RGB}{0,127,0}
\definecolor{mygreen}{RGB}{60,180,60}
\definecolor{lightblue}{RGB}{150,150,255}
\definecolor{blue}{RGB}{127,127,255}
\definecolor{darkblue}{RGB}{0,0,127}
\definecolor{myblue}{RGB}{80,80,200}
\definecolor{grey}{RGB}{127,127,127}
\definecolor{pink}{RGB}{255,180,180}
\definecolor{lightgrey}{RGB}{180,180,180}
\newcommand{\blue}[1][blue]{\textcolor{blue}{{#1}}}
\newcommand{\darkblue}[1][darkblue]{\textcolor{darkblue}{{#1}}}
\newcommand{\red}[1][red]{\textcolor{red}{{#1}}}
\newcommand{\orange}[1][orange]{\textcolor{orange}{{#1}}}
\newcommand{\darkgreen}[1][darkgreen]{\textcolor{darkgreen}{{#1}}}

\newcommand{\brown}[1][brown]{\textcolor{brown}{{#1}}}
\newcommand{\violet}[1][violet]{\textcolor{violet}{{#1}}}

\def\R{{\cal R}}

\def\X{{\cal X}}

\usepackage{verbatim}

\newcommand{\Sup}{{S}\xspace}

\newcommand{\SupB}{{S'}\xspace}

\newcommand{\SupC}{{\dot{S}}\xspace}
\newcommand{\SupCc}{{\dot{S}_c}\xspace}

\newcommand{\SupD}{{\mathring{S}}\xspace}
\newcommand{\SupDc}{{\mathring{S}_c}\xspace}

\newcommand{\support}{{support}\xspace}
\newcommand{\Support}{{Support}\xspace}

\newcommand{\measureC}{{optimality}\xspace}
\newcommand{\MeasureC}{{Optimality}\xspace}

\newcommand{\measureCab}{{opt}\xspace}

\newcommand{\measureCcab}{{opt_c}\xspace}

\newcommand{\permanence}{{permanence}\xspace}

\newcommand{\perm}{{perm}\xspace}
\newcommand{\permc}{{perm_c}\xspace}

\newcommand{\residual}{{residual}\xspace}

\newcommand{\resc}{{res_c}\xspace}

\usepackage{array}
\usepackage{algorithm} 
\usepackage{algorithmic}  

\newcommand{\arxiv}[1]{\textcolor{black}{{#1}}}
\newcommand{\gerl}{\violet[\sffamily{gErl}]\xspace}

\usepackage{amsmath}
\usepackage{amssymb}

\graphicspath{{}}

\newcommand{\qed}{\nobreak \ifvmode \relax \else
      \ifdim\lastskip<1.5em \hskip-\lastskip
      \hskip1.5em plus0em minus0.5em \fi \nobreak
      \vrule height0.75em width0.5em depth0.25em\fi}

\newcommand{\yaxis}{$y$-axis\xspace}
\begin{document}
%
\title{Forgetting and consolidation for incremental and cumulative knowledge acquisition systems}

\author{\small Fernando Mart{\mbox{\'{\i}}}nez-Plumed, C\`{e}sar Ferri, Jos$\acute{\mbox{e}}$ Hern$\acute{\mbox{a}}$ndez-Orallo, Mar${\mbox{\'{\i}}}$a Jos$\acute{\mbox{e}}$ Ram${\mbox{\'{\i}}}$rez-Quintana\\
\small DSIC, Universitat Polit\`ecnica de Val\`encia, Cam\'{\i} de Vera s/n, 46022 Val\`encia, Spain.\\
\small E-mails: {\tt \{fmartinez,cferri,jorallo,mramirez\}@dsic.upv.es}}

\maketitle

\begin{abstract}

The application of cognitive mechanisms to support knowledge acquisition is, from our point of view, crucial for 
making the resulting models coherent, efficient, credible, easy to use and understandable. In particular, there are two characteristic features of intelligence that are essential for knowledge development: forgetting and consolidation. Both plays an important role in knowledge bases and learning systems to avoid possible information overflow and redundancy, and in order to preserve and strengthen important or frequently used rules and remove (or forget) useless ones. We present an incremental, long-life view of knowledge acquisition which tries to improve task after task by determining what to keep, what to consolidate and what to forget, overcoming \emph{The Stability-Plasticity} dilemma \cite{grossSP}. In order to do that, we rate rules by introducing several metrics through the first adaptation, to our knowledge, of the Minimum Message Length (MML)  principle \cite{Wallace01081968} to a \emph{coverage graph}, a hierarchical assessment structure which treats evidence and rules in a unified way. The metrics are not only used to forget some of the worst rules, but also to set a consolidation process to promote those selected rules to the knowledge base, which is also mirrored by a demotion system. We evaluate the framework with a series of tasks in a chess rule learning domain.


$\:$

\end{abstract}



%

\section{Introduction}\label{sec:intro}


Machine learning and other data analysis techniques are becoming crucial for many applications where we want to turn (big) data into knowledge. However, any conception of knowledge discovery that aims at generating more insightful results must overhaul the whole process with an incremental, developmental perspective. The view cannot longer be a transformation from data to knowledge, but a transformation of knowledge (plus data) into new knowledge. As a result, properly representing, revising, evaluating, organising and retrieving previous knowledge is crucial in this quest for more complex, insightful, powerful and ultimately cognitive approaches to make knowledge discovery an incremental process.

Knowledge acquisition\footnote{In expert systems, the term knowledge acquisition is usually understood as the incorporation of expert knowledge into the system. In this paper, we use the term knowledge acquisition as the process of discovering new knowledge from facts and integrating it with the existing knowledge.}, understood as an automated process of abstracting knowledge from facts and other knowledge, cannot be understood as a naive accumulation of what is being learned. It should be checked whether new learned knowledge can be redundant, irrelevant or inconsistent with old one, and whether it may be built upon previously acquired knowledge. 
From our point of view, knowledge acquisition systems should be developed for this purpose. This lead us to one of the well-known constraints for AI systems: \emph{The Stability-Plasticity} dilemma \cite{grossSP}. The basic idea is that an AI system must be capable of learning new things (plasticity) without losing previously learned concepts (stability). 
This has been a designing principle mainly investigated within the perspective of neural computation over the last thirty years. Some of the proposed solutions include: (a) dual-memory systems simulating the presence of short and long-term memory \cite{Robins95catastrophicforgetting,French97pseudo-recurrentconnectionist,Ans1997989}, and (b) cognitive architectures such as the \emph{Adaptive Resonance Theory} (ART) \cite{Grossberg:2013:ART:2405841.2405958} emulating how the brain processes information. In both cases, catastrophic forgetting \footnote{Phenomenon by which neural networks completely forget previously learned information when exposed to new one.} of previously learned information was thereby effectively overcome, however, those approaches are only able to gain new knowledge (forgetting is not allowed) without proper management of existing knowledge,  
thus taking away versatility and efficiency to the proposals. 

From our point of view the above principle should point the way to a more general principle which also applies to general AI systems for knowledge acquisition. It could be used to define ``truly'' intelligent systems (a) able to support incrementally knowledge acquisition without the need to be discarded and retrained repeatedly (which is not cost-effective), (b) where the inductive and deductive reasoning algorithms are integrated for such a goal and guided by knowledge evaluation metrics, and, finally, (c) able to focus on what is relevant knowledge (or dually to discard what is not) by the use of cognitive mechanisms that simplify the learning of new knowledge. Following those requirements, below we overview some prior work in the area of knowledge acquisition.


Over the last decades, there has been an extensive work on growing knowledge bases from discovered patterns and rules. We find this in different areas, including expert systems, machine learning, cognitive science, nonmonotonic logic, information systems and inductive (logic) programming. 
For instance, \emph{Lifelong Machine Learning} (LML)\cite{Thrun96survey} is concerned with the persistent and cumulative nature of learning, namely: $(a)$ capable of retaining and using prior knowledge, and $(b)$ capable of acquiring new knowledge over a series of prediction tasks.
Similarly, \emph{Transfer Learning} \cite{Pan10surveyTL} and multitask learning \cite{Baxter00amodel, Caruana93multitasklearning} take a similar perspective, where it is more explicit that the process is task-oriented, and knowledge and its structure does not always play a central role in these systems. 
ELLA (Efficient Lifelong Learning Algorithm) \cite{ruvolo13} and NELL (Never-Ending Language Learner) \cite{Carlson10} are two more recent approaches to LML, which are able to integrate many capabilities. However, it is not easy to export or derive general principles from these works to analyse a knowledge base and help in a general incremental knowledge discovery process.

Other related topics are concept drift and theory revision  \cite{rendell1995,paes2005probabilistic,gama2014survey}, where some rules are replaced by new rules that are consistent with new experience. This is similar to the approach in nonmonotonic and approximate reasoning, and probabilistic or stochastic logic representations. 
The areas of inductive logic programming \cite{muggleton1994inductive,muggleton1999scientific} or general inductive programming \cite{flener2008introduction,gulwani2014inductive} have seen several approaches for incremental  \cite{ferri2001incremental} or cumulative systems \cite{henderson14}.




A crucial aspect relies on theory and knowledge evaluation. When the theory or hypothesis is considered as a whole and separated from the evidence, we have many well-founded proposal, such as the MML principle \cite{Wallace01081968,Wallace:2005:SII:1051763} or the similar (but posterior) MDL principle \cite{rissanen1999hypothesis,grunwald2005advances}. However, for knowledge integration and consolidation it is necessary to assess each part of the theory independently, where different parts of the theory can have different degrees of validity, probability or reinforcement \cite{hernandez2000constructive,hernandez2000explanatory}. However, there is still a separation between knowledge and evidence.  
It would be meaningful to provide a fully integration of knowledge and evidence into a hierarchical assessment structure from very specific and ground facts to more abstract rules. The perspective of a network or hierarchy of nodes that get support from other nodes is more common in the area of link analysis in web graphs such as the HITS algorithm \cite{Kleinberg:1999:HITS}, PageRank \cite{Brin:1998:ALH:297810.297827} or SALSA \cite{Lempel:2000:SAL:346241.346324}, or in infometrics. 



Finally, knowledge acquisition has much to learn from the study of human cognition \cite{Pedrycz1989305,Raducanu20081024,vandenBroek20081136,MartinezdelRincon20131849,Ramik20131577}. \arxiv{ We can fully realise the benefits of knowledge acquisition by paying attention to the cognitive factors that simplify the learning and processing of the knowledge which make the resulting models coherent, efficient, credible, easy to use and understandable \cite{pazzani2000knowledge}}. In particular, there is 
 a characteristic feature of intelligence that is essential for knowledge development: forgetting.  \arxiv{Meanwhile human memory has a positive connotation linked with performance, forgetting is often associated with negative terms as a state where memory does not work properly.}  Memory and forgetting are two complementary faces of the same biological process (synaptic plasticity), being the latter the one of the human mind's selective activities which allow us to abstract concepts.  It could be said that, without forgetting, memory would be completely useless. The absence of forgetting was masterly described  by Jose Luis Borges in his tale ``Funes, the Memorious'' (1942): ``To think is to forget a difference, to generalise, to abstract. In the overly replete world of Funes, there were nothing but details''. Clearly, remembering absolutely everything prevents from having abstract thought (the process of generalisation), given that induction and deduction rely on this ability. Therefore, in AI systems, forgetting should play an important role when acquiring knowledge. Forget has multiple shades of meaning in AI systems: it can refer to a complete and  irreversible elimination of significant old knowledge while learning new one;  or it can denote that 
new learned knowledge is not always kept in the working memory but abstractly encoded by identifying their relation to abstract concepts already present in the knowledge base. The first meaning clearly refers to those AI systems that are booted up for solving individual problems, whereas the latter definition is the desired one: forgetting should exist in knowledge bases and learning systems to avoid possible information overflow and redundancy, and in order to preserve and strengthen important or frequently used rules and remove (or forget) useless ones.

The ability to focus on what to discard what is not relevant is becoming more relevant not only in cognitive science and neuroscience \cite{quiroga2012nature}, but also in artificial intelligence (e.g., reasoning, planning, decision making). The notion of forgetting, also known as variable elimination, has been widely investigated in the context of classical logic (propositional and first-order logic) \cite{Lin94forgetit!,Lang03propositionalindependence,Lang:2010} and developed under the notion of logical equivalence, that is, logically equivalent formulas (theories) will remain equivalent after forgetting the same set of propositional variables or literals. 
A similar approach but for reasoning from inconsistent propositional bases is proposed in \cite{Lang:2010}. Recently, the concept of forgetting has been widespread in other non-classical logic systems from various perspectives such as in logic programs \cite{Zhang2006739,Eiter20081644, wang2012forgetting} where  
a semantic forgetting is used instead of developing a number of criteria for forgetting atoms; in modal logic \cite{Zhang20091525,su2009variable,liu2011progression} where variable elimination is applied in the context of intelligent agents; and in description logic (DLs) \cite{Wang2010} for omitting concepts and roles in knowledge bases. Forgetting (abstracting from) actions in planning has been also investigated in \cite{ErdemF07}. Finally, in \cite{zhao2012online} is proposed a forgetting mechanism for an online learning algorithm to learn sequential data with timelines able to gradually expel the outdated data that could become a possible source of misleading information.

From our point of view, forgetting in a knowledge base is closely linked to the previous concept of theory and knowledge evaluation. 
 Therefore, inspired by the MML principle, the informativeness of a piece of knowledge (in terms of usefulness or the opposite concept, irrelevance) can be assessed quantitatively only by its relationship between complexity and compression. This lead us to an easy and general concept of forgetting where as much information as possible from the original knowledge is preserved,  thus setting aside tasks such as the preservation of logical equivalences or the satisfaction of semantic properties between theories. 

Closely related with the above concept we found \emph{memory consolidation}, namely, the neurological process of converting information from short-term memory into long-term memory. Some studies about episodic memory in humans \cite{Shastri2001,tagkey2008iii} claim that memory traces in the hippocampus are not permanent and are occasionally transferred to neocortical areas in the brain through a consolidation processes. This consolidation process refers to the idea that memories continue to strengthen after they have been formed in the human brain and seems a primary factor underpinning memory and forgetting in knowledge bases and learning systems. Notwithstanding a single recent cognitive model of memory ascribes too much importance to consolidation procedures \cite{della2010forgetting}, we consider that not only forgetting must be a prevalent operation in knowledge acquisition, but also consolidating is crucial as well for promoting efficient memory storage.

Given the above overview, we see that it is not easy to develop a new knowledge discovery system that is meant to be cumulative. In fact, this research started when developing our system \gerl \cite{gerlNFMCP12,ICMLA2014}. We were looking on a proper foundation for detailed knowledge assessment metrics and criteria for forgetting. The need of making general principles available for our system and other systems motivated the current work.

\begin{figure}[!htbp]

\begin{center}
\includegraphics[width=0.5\columnwidth]{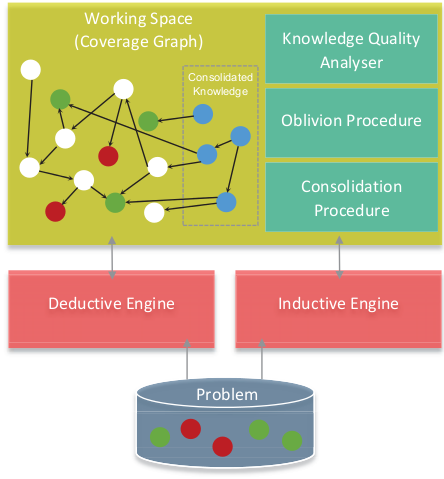} 
\end{center}
\caption{\arxiv{Architecture of a long-life knowledge discovery approach.}
}
\label{fig:system}

\end{figure}

In this work we take a most general approach 
 by considering that we start with an off-the-shelf {\em inductive engine} (e.g., a rule learner, an inductive logic programming (ILP) system \cite{muggleton1994inductive,muggleton1999scientific} or an inductive programming (IP) system \cite{flener2008introduction,gulwani2014inductive}) and an off-the-shelf {\em deductive engine} (e.g., a coverage checker, an automated deduction system or a declarative programming language) 
 and, over them, we build an long-life knowledge discovery system (see Figure \ref{fig:system}).

For this purpose, several issues have to be addressed:
\begin{enumerate}
\item The inductive engine can generate many possible hypotheses and patterns. Once brought to working memory we require metrics to evaluate how these hypotheses behave and how they are related in the context of previous knowledge.  Additionally, at any time new evidence can be added as rules to the working space.
\item  As working memory and computational time are limited, we need a forgetting criterion to discard some rules which are considered irrelevant in terms of informativeness.  
\item  The deductive engine checks the coverage of each hypothesis independently, using the background or consolidated knowledge as auxiliary rules,  but not other working rules. As a result, only when new knowledge is consolidated we can use it for new problems or for more difficult examples of the same problem.  This means that deduction is ``modulo the background knowledge''. In other words, working hypotheses must be able to use consolidated knowledge but not other working rules. 
\item The promotion of rules into consolidated knowledge must avoid unnecessarily large knowledge bases and  the consolidation of rules that are useless, too preliminary or inconsistent.  This means that rules must promoted and demoted.  Also, if the knowledge base becomes too large, finding the appropriate pieces of knowledge for new tasks will be less efficient. This means that rules must promoted and demoted to keep a powerful, but still manageable knowledge base.
\end{enumerate}

\noindent The idea of coverage graph is used as the basis for structuring knowledge and is delegated to the deductive engine. The generation of new rules is delegated to the inductive engine. The crucial part is the definition of appropriate metrics to guide the way knowledge develops. For this purpose, the MML principle is used as a sound theoretical ground for the metrics. 

The paper is organised as follows. Section \ref{sec:coverage} introduces the notion of coverage graph, which is our setting for a knowledge base. 
Over this coverage graph, we are able to introduce an adaptation of the MML principle and related metrics in section \ref{sec:basic}.
Section \ref{sec:structuring} deals with knowledge structuring, how rules are forgot, promoted and demoted. 
We include several experiments where we illustrate how knowledge consolidation and forgetting works in section \ref{sec:experimental}. Finally, section \ref{sec:conclusions} closes the paper with the contributions and some future work.

\section{Coverage graph}\label{sec:coverage}

We consider that `rules' are used for expressing examples, hypotheses and background knowledge. Rules are denoted as $e$ where $class(e) = c$,  $c \in C$ and $C$ is the set of classes, such as $\{ \mbox{false}, \mbox{true}\}$. 
The set of all possible rules is denoted by $\R$, where $W \subset \R$  is the working space or memory, and $K \subset \R$ is the background or consolidated knowledge base. 

Rules are presented as vertexes or nodes $V$  (and we refer them indistinctly) in a directed acyclic graph $G(V,A)$ we call {\em coverage graph} (which is the DAG representation of a specific working space), because the directed edges $A$ represent the coverage relation between the different rules  
as determined by the deductive engine. 
We say that a rule $\rho_a$ is covered by another rule $\rho_b$ if $ (K \cup \rho_b) \models \rho_a$. The precise understanding of the semantic consequence operator will depend on the rule representation language used and the deductive engine. Hence, if there is an edge $a=(\mu,\nu)$ (or $\mu \rightarrow \nu$), then $\nu$ is said to be directly covered by $\mu$ using $K$\footnote{For simplicity, the \emph{coverage graphs} do not include the edges for the transitive closure of the covering relation, i.e., if a node $\mu$ covers nodes $\nu$ and $\gamma$, but $\nu$ also covers $\gamma$, only the edges $\mu \rightarrow \nu$ and $\nu \rightarrow \gamma$ are included in the graph.}.

The set of ancestors  and successors of a node $\nu$ are defined as $anc(\nu) =  \{\mu | \mu \rightarrow \nu \}$ and $suc(\mu) =  \{\nu | \mu \rightarrow \nu \}$ (respectively). Also, we distinguish two subsets of nodes: $leaves$, nodes without successors ($|suc(\nu)| = 0$), where the set of $leaves$ $\nu$ of class $c$ is denoted as $leaves_c$; and $roots$, nodes without ancestors ($|anc(\nu)|=0$).

\begin{figure}[!htbp]
\begin{center}
\includegraphics[width=0.45\columnwidth]{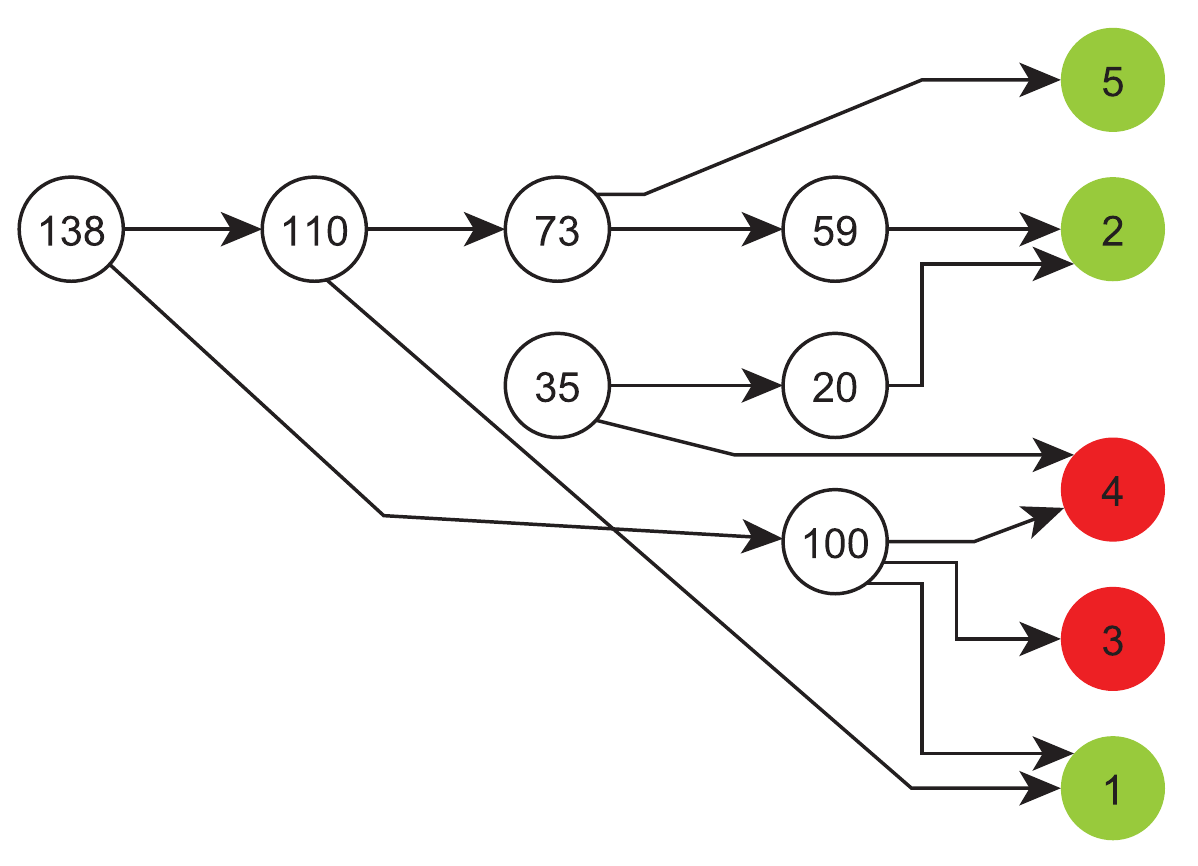} 
\end{center}

\vspace{-0.3cm}

\caption{\emph{Coverage Graph} of the \emph{family relations} problem. Green and red nodes refer to positive and negative examples respectively. The graph shows rule IDs according to Table \ref{tab:tablefamily}. }
\label{fig:graphFamily}
\end{figure}

Figure \ref{fig:graphFamily} shows an example of \emph{Coverage Graph} 
 of a well-known ILP problem \cite{muggleton1994inductive}: the family relationship. In this problem, the task is to define the target relation $daughter(X,Y)$, which states that person $X$ is daughter of person $Y$. $W$ consists of three positive examples (rules $1$, $2$ and $5$), two negative ones (rules $3$ and $4$), and seven selected rules that try to generalise and solve the problem (Table \ref{tab:tablefamily} right), whereas $K$ is composed of the relations \emph{female} and \emph{parent} (Table \ref{tab:tablefamily} left). Note that the rules in $K$ have not been included in the graph for clarity, although they belong to the initial ``consolidated knowledge''. 

\begin{table}[!htbp]
\begin{center}
{\scriptsize
\begin{tabular}{ll|ll}\hline
\multicolumn{2}{c|}{{\bf Background Knowledge}}&\multicolumn{2}{c}{{\bf Rules}}\\
\hline
ID &Rule&ID &Rule \\
\hline
k1&parent(ann, mary).&1&daughter(mary,ann).\\
k2&parent(ann, tom).&2&daughter(eve,tom).\\
k3&parent(tom, eve).&3&daughter(tom,ann).\\
k4&parent(tom, ian).&4&daughter(eve,ann).\\
k5&female(ann).&5&daughter(cris,tom).\\
k6&female(mary).&100&daughter(X,Y):- female(Y),parent(Y,mary).\\
k7&female(eve).&59&daughter(eve,tom):-  female(eve),parent(tom,eve).\\
&&20&daughter(eve,tom):-  female(eve).\\
&&35&daughter(eve,Y):-  female(eve).\\
&&73&daughter(X,tom):-  female(X),parent(tom,X).\\
&&110&daughter(X,Y):-  female(X),parent(Y,X).\\
&&138&daughter(V,W):-  female(X),parent(Y,Z).\\
\end{tabular}
}
\end{center}

\caption{Left: Background Knowledge for  the \emph{family relations} problem. Right: Rules of this problem in {\sf Prolog} notation.}
\label{tab:tablefamily}

\end{table}%

\section{Basic Metrics for Discovered Knowledge Assessment}\label{sec:basic}

In order to select and arrange the set of rules in the working space, various measures of usefulness, relevance and consistency have to be derived from the \emph{coverage graph}. Based on the idea that the relevance or usefulness of a rule can be stated by the relationship between its own complexity and the complexity of the rules it covers, a general criterion such as the \emph{Minimum Message Length} \cite{Wallace01081968} (MML) can be used as a starting criterion from which to derive new metrics.


\subsection{Minimum Message Length}

The \emph{Minimum Message Length} is one of the most popular selection criterion in inductive inference (for a formal justification and its relation to Kolmogorov complexity and the related MDL principle, see  \cite{Li:2008:IKC:1478784, DBLP:journals/cj/WallaceD99a,Wallace:2005:SII:1051763}).  It provides an interpretation of the Occam's Razor principle: the model generating the shortest overall message (composed by the model and the evidence concisely encoded using it) is more likely to be correct. This message can be re-stated  in a Bayesian form \cite{Wallace01081968} with the length of the first part of the message (the model) and the length of the second part (evidence covered). The Bayesian theorem, which is the primary concern of Bayesian inference, is shown in equation \ref{bayesianMML}:

\begin{equation} \label{bayesianMML}
P(H|E) =\frac{P(H) \cdot P(E|H)}{P(E)}=\frac{P(H \cap E)}{P(E)}
\end{equation}

\noindent where $P(H)$ is the prior probability of the model $H$, $P(E|H)$ is the likelihood, and $P(E)$ is the probability of the evidence $E$. An information-theoretic interpretation of MML is that a given evidence $E$ of probability $P(E)$ can be coded by a message of length $L(E)= -log_2(P(E))$ \cite{citeulike:1584479}. Therefore, taking the negative logarithm of the expression \ref{bayesianMML} and according to the MML philosophy, the length of a hypothesis $H$ given a fixed evidence $E$ $(L(H|E)$ is defined as the sum of three simple heuristics: a complexity-based heuristic (which measures the complexity of $H$), a coverage heuristic (which measures how much extra information is necessary to express the evidence given the hypothesis $H$) and the length of the evidence ($L(E)$) which equal for all competing hypotheses: 

\begin{equation} \label{sizeMML}
L(H|E)= L(H) + L(E|H) - L(E)
\end{equation}

By minimising equation \ref{sizeMML}  we maximise the posterior probability. This involves searching for the model that gives the shortest message. 


Apart from its connection with Kolmogorov complexity and Solomonoff induction \cite{LiVitanyi}, which gives additional support for its use, the MML principle (and the similar MDL principle) has been successfully applied in many areas of machine learning, AI and cognitive science. However, to our knowledge, the MML principle has always been applied to select between hypotheses with respect to some given evidence. In our case, we have a coverage graph where rules cover other rules, so they become $H$ and $E$ at the same time. In a way, what we need is a hierarchical MML application, with this in mind 
the MML principle can be adapted to be used in our approach with the following considerations: instead of measuring the length of a hypothesis $H$ given fixed evidence $E$, what we want to measure is the length of each rule $\rho$ in $W$ with respect to the rest of rules in $W$ (which includes examples and hypotheses) because $\rho$ can model not only examples, but also other rules. Therefore, $L(\rho| W)$ is defined as the sum of the length of $\rho$ ($L(\rho)$), and the length necessary to express the rules in $\{W - \rho\}$ not modelled by $\rho$ ($L(W|\rho)$), minus the length of the total rules in $W$ ($L(W)$). Formally:

\begin{equation} \label{sizeMMLgraph}
L(\rho|W)= L(\rho) + L(W|\rho) - L(W)
\end{equation}

Apparently, it just seems a notational change wrt. Eq.~\ref{sizeMML}. This is only true for the first term, which is estimated in the same way as the original MML principle. The term $L(\rho)$ can be defined in different ways depending on the rule representation language. For instance, if we are using logical or functional rules  (as in the family example), we could use the following approximation. Given $\Sigma$ a set of $m_{\Sigma}$ functor symbols of arity $\ge 0$, and $\X$ a set of $m_{\X}$ variables, we could define the length of a rule $\rho$ containing $n_{\Sigma}$ functors and $n_{\X}$ variables as 

\begin{equation}\label{msglen}
\begin{split}
L(\rho) & \triangleq  m_{\Sigma} \log_2(n_{\Sigma}+1)\\
 &+ \frac{m_{\X}}{2} \log_2(n_{\X}+1)
\end{split}
\end{equation}

\noindent Note that we promote variables over constants or functors.

Table \ref{tab:tablefamilyMetrics} shows the length in bits and the class for the rules in the graph of Figure \ref{fig:graphFamily}.

\begin{table}[!htbp]
	
\begin{center}

\begin{tabular}{r|r|c}
{\bf ID} &{\bf $L(\rho)$}& {\bf class}\\ \hline
1&17.844&$+$\\ 
2&17.844&$+$\\ 
3&17.844&$-$\\ 
4&17.844&$-$\\ 
5&17.844&$+$\\ 
100&11.977&\\ 
59&20.036&\\ 
20&11.591&\\ 
35&9.284&\\ 
73&13.114&\\ 
110&9.962&\\ 
138&12.462&\\ 
\end{tabular}

\end{center}

\caption{Length and class for the rules on the right side of Table \ref{tab:tablefamily}.}
\label{tab:tablefamilyMetrics}
\end{table}%


\subsection{MML goes hierarchical: \Support}

Following with the equation \ref{sizeMMLgraph}, we are going to reunderstand the terms $L(W|\rho) - L(w)$ to be adapted to \emph{coverage graphs} and multiclass settings. Roughly speaking, these terms capture the ``net profit'' of the rules both in terms of \support or coverage (length in bits of the rules covered). More formally, we define the \support of a rule $\rho \in W$ as:


\begin{equation} \label{gain1}
\Sup(\rho, W) \triangleq L(\rho) - L(\rho|W) = L(W) - L(W|\rho)
\end{equation}

\noindent where $L(W) - L(W|\rho)$ represents the coverage of a rule $\rho$ expressed in bits, that is, the length of all the rules in $W$ minus the length of the rules not covered by $\rho$. Therefore, the \support of a rule $\rho$ represents the length of the rules it covers: 

\begin{equation} \label{Support}
\Sup(\rho,W) = \sum\limits_{\nu: \rho \models \nu}L(\nu) 
\end{equation}

\noindent 
leading to an alternative expression for $L(\rho|W)$ (eq.~\ref{sizeMMLgraph}) in terms of \support:

\begin{equation} \label{L1}
L( \rho | W) = -\Sup(\rho, W) + L(\rho)
\end{equation}

\noindent which establishes that maximising $\Sup(\rho,W)$ and minimising $L(\rho)$ we minimise $L( \rho | W)$ which involves searching for the rule $\rho$ that covers the maximum number of rules and has the lowest length.

The following step is to adapt eq.~\ref{L1} to be used in \emph{coverage graphs} 
that does not explicitly include the edges for the transitivity of the coverage relation.
In order to consider the upwards propagation, only the $leaves$ will have an initial \support value which is equal to its length in bits, and the rest of nodes will distribute it recursively by propagating this \support. 
 Thus,  the new \support ($\SupB(\rho,W)$) adapted to work on \emph{coverage graphs} is defined as:

\begin{equation} \label{Support2}
\SupB(\rho,W) \triangleq 
\begin{cases}
    L(\rho) & \text{if } \rho \in leaves \\
    \sum\limits_{\nu \in suc(\rho)}\SupB(\nu,W)& \text{otherwise}
\end{cases}
\end{equation}

In order to avoid the scenario where the less grounded (upper) nodes get higher and higher \support values, the \support measure is required to satisfy a \emph{conservative} condition. This property is somehow related to the law of conservation of energy, implying that at any node in a coverage graph, the sum of the total \support flowing into that node is equal to the sum of the total \support flowing out of that node. 


%
Now, to make $\SupB$ conservative we need to divide the \support coming from the  outcoming of a specific node $\nu$ by $|anc(\nu)|$ in order to equally distribute the support of  $\nu$ between all of its ancestors. 

Therefore, the new formula used to calculate the \support of a rule ($\SupC(\rho,W)$) is defined to be equal to:

\begin{equation} \label{Support3}
\SupC(\rho,W) \triangleq 
\begin{cases}
    L(\rho) & \text{if } \rho \in leaves\\
              
    \sum\limits_{\nu \in suc(\rho)}\frac{\SupC(\nu,W)}{|anc(\nu)|} & \text{otherwise}

\end{cases}
\end{equation}

\noindent and leading to an expression for $L(\rho | W)$ (\ref{L1}) in terms of this conservative \support:

\begin{equation} \label{L2}
\dot{L}(\rho | W) = -\SupC(\rho,W) + L(\rho)
\end{equation}

\begin{figure}[!htbp]
\begin{center}
\includegraphics[width=0.5\columnwidth]{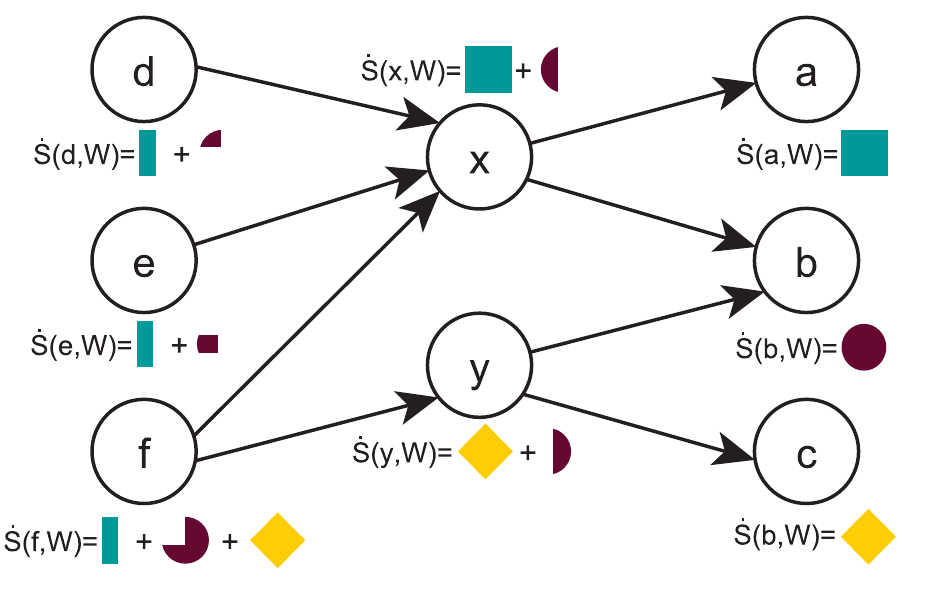} 
\end{center}
\vspace{-0.5cm}
\caption{Graphical representation of the flow of the \support by using equation \ref{Support3} through the \emph{coverage graph}: the \support of each \emph{leave} node is always allocated in the \emph{roots}. Therefore, the total \support of the \emph{leaves} is equal to the total \support in the \emph{roots}}
\label{fig:cons2}
\end{figure}

Equation \ref{Support3} now accomplishes the mandatory conservative condition which could be stated such as the \support of a node (which depends on its successors) has to be always entirely allocated in its ancestors together with the \support inherited from other covered nodes (see Figure \ref{fig:cons2}).






This implies (but not vice versa) that the total sum of the \support in the \emph{leaves} in the \emph{coverage graph} is equal to the total sum of the \support at the \emph{root} nodes. Namely: 


\begin{equation} \label{dem2}
\sum\limits_{\mu \in leaves}\SupC(\mu,W) = \sum\limits_{\nu \in roots} \SupC(\nu,W)
\end{equation}



\arxiv{For each \emph{leaf} in the \emph{coverage graph} we have $n$ different paths whereby the support flows upwards to \emph{root} nodes. Whenever a path is forked (an ancestor is found), the \support is always divided by the number of the outcoming paths, having the ancestors an equally part of the \support and thus having the roots a proportion of the original support of the leaves transitively covered by them. Therefore, if we assume that the total \support at the roots is different from the total \support at the \emph{leaves}, it means that an external transfer of \support (which comes from or goes to other sources) has happened. However, accordingly to eq.~\ref{Support3}, this is not possible and, therefore, the total sum of the \support at the \emph{roots} always remains constant and equal to the total \support at the \emph{leaf} nodes (see Figure \ref{fig:cons2}).}

\begin{bclogo}[logo = \bccrayon]{Example}
{\footnotesize
\arxiv{Viewed through the example in Figure \ref{fig:cons2} and accordingly to the equation \ref{Support3} we have that the \support of the \emph{root} nodes is}

\begin{equation*} 
\SupC(d,W)  = \SupC(e,W) = \frac{\SupC(x,W)}{3},  
\SupC(f,W)  = \frac{\SupC(x,W)}{3} + \SupC(Y,W),
\end{equation*}

\noindent where

\begin{equation*} 
\SupC(x,W)  = \SupC(a,W) + \frac{\SupC(b,W)}{2},
\SupC(y,W)  = \frac{\SupC(b,W)}{2} + \SupC(c),  
\end{equation*}

\noindent and

\begin{equation*} 
\SupC(a,W)  = L(a),
\SupC(b,W)  = L(b),
\SupC(c,W)  = L(c)
\end{equation*}

\noindent thus making the following equations true (accordingly to the formula \ref{Support3}):

\begin{equation*} 
\begin{split}
\underbrace{\SupC(a,W) + \frac{\SupC(b,W)}{2}}_{\SupC(x,W)} - \underbrace{\SupC(d,W)}_{\frac{\SupC(x,W)}{3}} - \underbrace{\SupC(e,W)}_{\frac{\SupC(x,W)}{3}} - \underbrace{(\SupC(f,W) -  \frac{\SupC(y,W)}{2})}_{\frac{\SupC(x,W)}{3}} = 0 \\
\underbrace{\SupC(c,W) + \frac{\SupC(b,W)}{2}}_{\SupC(y,W)} - \underbrace{(\SupC(f,W) - \frac{\SupC(x,W)}{3})}_{\SupC(y,W)} = 0
\end{split}
\end{equation*}

\noindent and, also, being the total \support at leaf nodes ($L(a) + L(b) + L(c)$) equal to the total \support at root nodes (equation \ref{dem2}):

\begin{equation*} 
\begin{split}
 \SupC(d,W) + \SupC(e,W) +  \SupC(f,W) & = (\frac{\SupC(x,W)}{3}) +  (\frac{\SupC(x,W)}{3}) + ( \frac{\SupC(x,W)}{3} +  \SupC(y,W))\\
 & = \Sup(x,W) + \SupC(y,W)\\
 & = ( \SupC(a,W) + \frac{\SupC(b,W)}{2} ) + ( \frac{\SupC(b,W)}{2} + \SupC(c,W)  )\\
 & = \SupC(a,W) + \SupC(b,W) + \SupC(c,W)
\end{split}
\end{equation*}
}
\end{bclogo}

Finally, we need to take into account that, since the working space $W$ can accommodate examples of different classes, we need our metric to distinguish between them and, hence, there are as many \support values for each node as many different classes there are in the working space, each one holding the \emph{conservative} property and formally defined to be equal to: 

\begin{equation} \label{Support4}
\SupCc(\rho,W) \triangleq
\begin{cases}
    L(\rho), & \text{if } \rho \in leaves_c\\
    \sum\limits_{\nu \in suc(\rho)}\frac{\SupCc(\nu,W)}{|anc(\nu)|} & \text{otherwise}
\end{cases}
\end{equation}

\noindent with eq.~\ref{L2} being defined for classes as follow:

\begin{equation} \label{gain3}
-\dot{L}_c(\rho | W) = \SupCc(\rho,W) - L(\rho) 
\end{equation}

\noindent The value of $\dot{L}_c$ is interpreted as the hierarchical version of the MML principle, with $\dot{L}_c$ being the lower the better (and obviously $-\dot{L}_c$ the higher the better).

Following with the \emph{Family} example, Table \ref{tab:tablefamilyMetrics1} shows the \support and the negative form of  $L(\rho| W)$  (for each class) of the rules in the graph in Figure \ref{fig:graphFamily}.

\begin{table}[!htbp]
	
\begin{center}

\begin{tabular}{r|r|c|r|r|r|r}
{\bf ID} &{\bf $L(\rho)$}& {\bf class} &{\bf $\SupC_+$} &{\bf $\SupC_-$} &{\bf $-\dot{L}_{+}$} &{\bf $-\dot{L}_{-}$} \\ \hline
1   &17.844     &$+$      &17.844     &0.0      &0.0        &-17.844         \\ 
2   &17.844     &$+$      &17.844     &0.0      &0.0        &-17.844         \\ 
3   &17.844     &$-$      &0.0        &17.844   &-17.844    &0.0        \\ 
4   &17.844     &$-$      &0.0        &17.844   &-17.844    &0.0       \\ 
5   &17.844     &$+$      &17.844     &0.0     	&0.0        &-17.844         \\ 

100 &11.977     &       &8.922      &26.766   &-3.0549    &14.788    \\
59  &18.791     &       &8.922      &0.0      &-11.114    &-20.036         \\ 
20  &11.591     &       &8.922      &0.0      &-2.668     &-11.591         \\ 
35  &9.284      &       &8.922      &8.922    &-0.362     &-0.362    \\ 
73  &13.114     &       &26.766     &0.0      &13.651     &-13.114       \\ 
110 &9.962      &       &35.688     &0.0      &25.726     &-9.962    \\ 
138 &12.462     &       &44.61      &26.766   &32.147     &14.303    \\ 
\end{tabular}

\end{center}

\caption{$\SupC$ and $-\dot{L}(\rho| W)$ values (both for the $+$ and $-$ classes) for the rules on the right side of Table \ref{tab:tablefamily}. Taking a look at the table, we cannot decide which is the best rule in global terms: we can only establish a ranking per classes (by using the \support values) without taking into account any other information. 
}
\label{tab:tablefamilyMetrics1}
\end{table}%


\subsection{\MeasureC}



By using the \support as the sole criterion to rank the rules in $W$ is useful provided there are only rules belonging to one class. However, when there are more than one class in $W$, we need to consider the purity or confidence of the rules. 
In the same spirit of the MML principle, 
we define the \measureC as the difference between the cost of coding a rule following equation \ref{gain3} for a specific class and  the cost of coding the exceptions, i.e.,: the \support of the rules covered that belong to the other classes. 
We use a factor $\beta$  
 indicating the relevance of rules being as pure as possible. 
Formally:

\begin{equation} \label{opt1}
\measureCcab(\rho,W) \triangleq   -\beta \cdot \dot{L}_c(\rho | W)  - (1 - \beta) \cdot \sum_{\substack{c' \in C \\ c' \neq c}} \SupC_{c'}(\rho,W)
\end{equation}

\noindent leading to a {\em generic} 
 \measureC of a rule as:

\begin{equation} \label{opt2}
\measureCab(\rho,W) \triangleq  \max\limits_{c \in C} (\measureCcab(\rho,W))
\end{equation}

\noindent Following with the \emph{Family} example, Table \ref{tab:tablefamilyMetrics3}  shows the \measureC values per class (the generic \measureC in bold) for the rules in the graph of Figure \ref{fig:graphFamily} using $\beta=0.5$.  
According to these values, rule $110$ is the most significant rule, as it can be easily viewed in the \emph{coverage graph} because it covers all the positive examples and no negative one.

\begin{table}[!htbp]
	
\begin{center}

\begin{tabular}{r|r|c|r|r|r|r|r|r}
{\bf ID} &{\bf $L(\rho)$}& {\bf class} &{\bf $\SupC_+$} &{\bf $\SupC_-$} &{\bf $-\dot{L}_{+}$} &{\bf $-\dot{L}_{-}$} &{\bf $\measureCab_+$} &{\bf $\measureCab_-$}\\ \hline

1   &17.844     &$+$      &17.844     &0.0      &0.0        &-17.844     &{\bf 0.0}         &-17.844           \\ 
2   &17.844     &$+$      &17.844     &0.0      &0.0        &-17.844     &{\bf 0.0}         &-17.844           \\   
3   &17.844     &$-$      &0.0        &17.844   &-17.844    &0.0         &-17.844           &{\bf 0.0}   \\ 
4   &17.844     &$-$      &0.0        &17.844   &-17.844    &0.0         &-17.844           &{\bf 0.0}   \\
5   &17.844     &$+$      &17.844     &0.0     	&0.0        &            &0.0               &-17.844           \\ 

100 &11.977     &       &8.922      &26.766   &-3.0549    &14.788      &-14.91			  	  &{\bf 2.933} \\
59  &18.791     &       &8.922      &0.0      &-11.114    &-20.036     &{\bf -5.557}      &-14.479           \\   
20  &11.591     &       &8.922      &0.0      &-2.668     &-11.591     &{\bf -1.334}      &-10.256           \\     
35  &9.284      &       &8.922      &8.922    &-0.362     &-0.362      &{\bf -4.642}      &{\bf -4.642}   \\ 
73  &13.114     &       &26.766     &0.0      &13.651     &-13.114     &{\bf 6.825}       &-19.939           \\   
110 &9.962      &       &35.688     &0.0      &25.726     &-9.962      &{\bf 12.863}      &-22.825           \\
138 &12.462     &       &44.61      &26.766   &32.147     &14.303      &{\bf 2.69}        &-15.153 \\

\end{tabular}

\end{center}

\caption{\MeasureC values (both for the $+$ and $-$ classes) for the rules on the right side of Table \ref{tab:tablefamily}. Bold values indicates the generic \measureC (equation \ref{opt2}). 
Ranking the rules by \measureC we see that the best rule is $110$. 
}
\label{tab:tablefamilyMetrics3}
\end{table}%




\section{Structuring knowledge: forgetting, promotion and demotion}\label{sec:structuring}

In our setting, rules are repeatedly generated by the inductive engine and added to the working space $W$. As an answer to the possible never-ending growth of $W$, 
it is necessary to 
have mechanisms for forgetting or revising useless pieces of acquired knowledge.  Using the metrics we have just introduced, we need a mechanism to discard those rules that are not useful, are inconsistent or do not get enough \support.

\subsection{Forgetting mechanism}

The \measureC of a rule $\rho$ is a core metric to determine its usefulness, but it is also important to see whether $\rho$ could be considered superfluous because it is covered (transitive or directly) by another rule of higher \measureC. If it is the case, $\rho$ is mostly redundant and it could be discarded safely. This idea leads to the following definition for the \permanence of a rule: 
\begin{equation}\label{perm}
\permc(\rho,W) \triangleq   \measureCcab(\rho) - \max(0,\max\limits_{\nu: \nu \models \rho} \measureCcab(\nu))
\end{equation}



\noindent with a {\em generic}  permanence: 

\begin{equation}\label{permGlobal}
\perm(\rho,W) \triangleq \max\limits_{c \in C} (\permc(\rho,W))
\end{equation}


\noindent 
The lower the value of permanence a rule has, the higher the odds it has to be forgotten.


\begin{figure}[!htbp]
\begin{center}
\includegraphics[width=0.8\columnwidth]{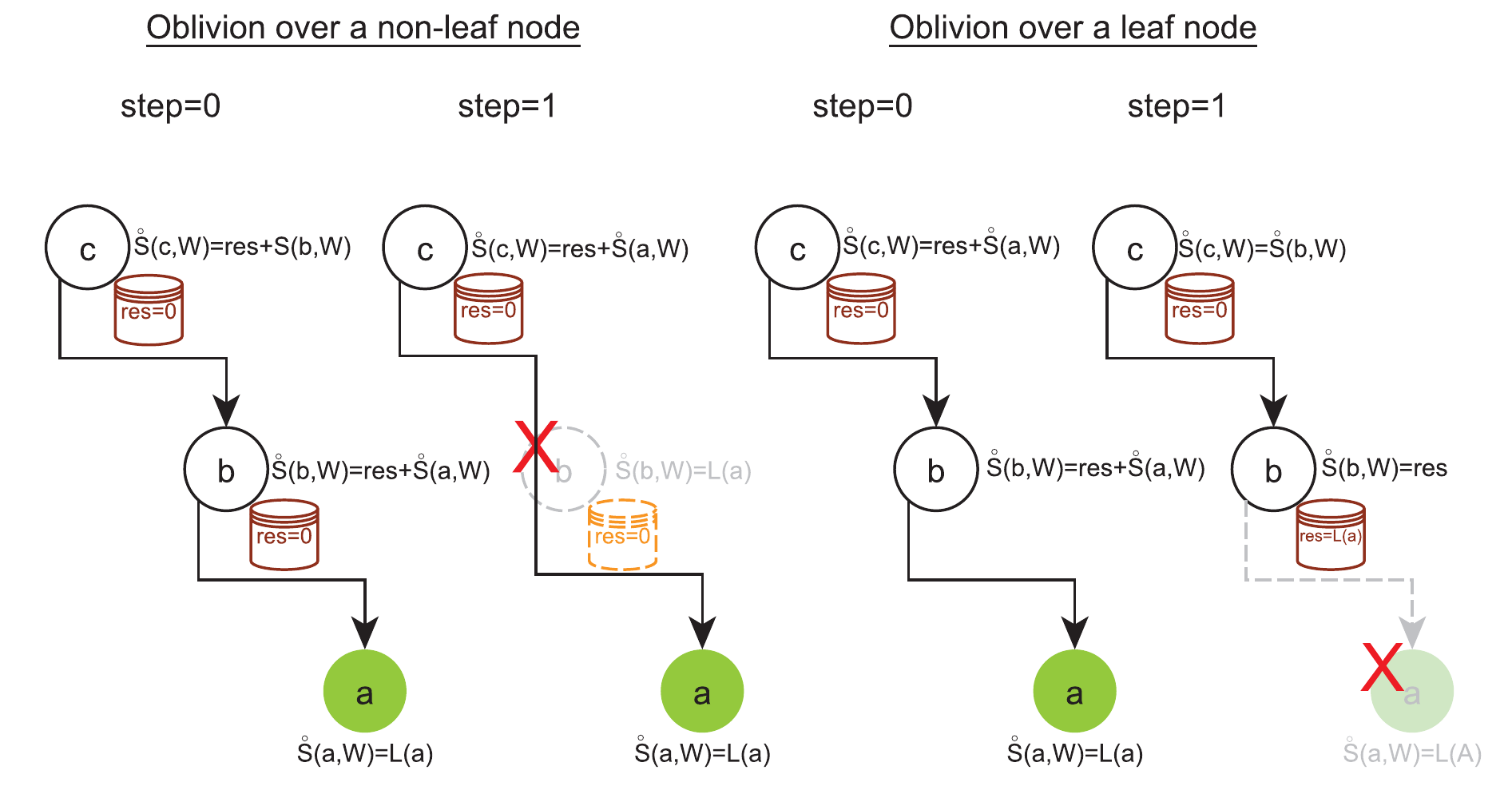} 
\end{center}

\caption{(Left) forgetting case $(a)$: an internal node is forgotten (graphically represented with a red cross). (Right) forgetting case $(b)$: a \emph{leaf} node is forgotten. 
Brown data storage cylinders graphically represent the concept of ``\residual'' that collects the \support  (for each class) of forgotten \emph{leaf} nodes. }
\label{fig:ob1}
\end{figure}

When we perform a forgetting step, the coverage graph is affected and coverages are also affected. 
%
In order to keep as much information about the past support, each rule is provided with a trace of its old \support. 
In cognitive systems this is associated to notions such as the preservation of belief and trust even if we forget the particular cases that gave support to a given statement. 
Therefore, the forgetting mechanism will work as follows:

\begin{enumerate}
\item If a non-\emph{leaf} node is selected to be forgotten, the \support of its successors has to be re-distributed among their ancestors and the ancestors of the forgotten node (see Figure \ref{fig:ob1} (left)).
\item In case there is a forgetting step that removes a leaf node, its \support 
 has to be equally distributed among the rules that cover it which inherit it as their  ``\residual''  \support value associated to each class $c$ ($\resc$) (see Figure \ref{fig:ob1} (right)).
\end{enumerate}

\begin{figure}[!htbp]
\begin{center}
\includegraphics[width=0.9\columnwidth]{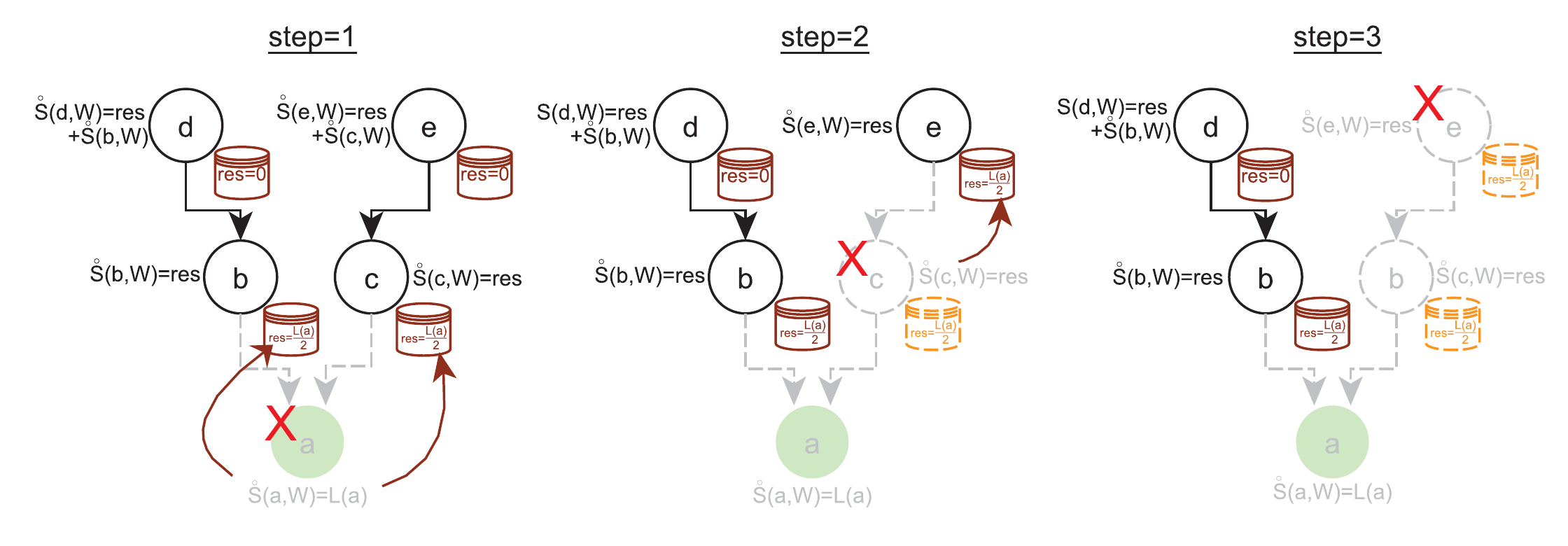} 
\end{center}
\caption{Forgetting mechanism performed over a complete branch. The conservative property over the \support measure occurs in all steps, but the initial amount of \support  at $step=0$ ($\sum\limits_{\mu\in leaves} \SupD(\mu,W) =  L(A)$) has been reduced at the last $step=4$  ($\sum\limits_{\mu\in leaves} \SupD(\mu,W) = \frac{L(A)}{2}$) due to the forgetting mechanism. }
\label{fig:ob2}
\end{figure}

Hence, the equation \ref{Support4} is modified to include the residual:

\begin{equation} \label{gain6}
\SupDc(\rho | W) \triangleq  
\begin{cases}
    L(\rho) & \text{if } \rho \in leaves_c \\
              
    \resc + \sum\limits_{v \in suc(\rho)}\frac{\SupDc(v,W)}{|anc(v)|} & \text{otherwise}
  
\end{cases}
\end{equation}

\noindent where $\resc$ is initially set as $0$. 
For each forgetting step, the \support of forgotten nodes is distributed among the outcoming nodes increasing their $\resc$ value, but if the last forgetting step removes a node without ancestor nor successors and a non-zero $res_c$, this value cannot be distributed and, therefore, is lost. These results in a decrease of the total \support of the graph: although the \support will remain conservative, the total amount will be lower than the  total \support of the \emph{coverage graph} before the forgetting steps. Consequently, in the end some rules may have an under-estimated \support value in terms of how many rules (of different classes) cover (see Figure \ref{fig:ob2}).

\begin{figure*}[!htbp]
\begin{center}
\includegraphics[width=0.8\textwidth]{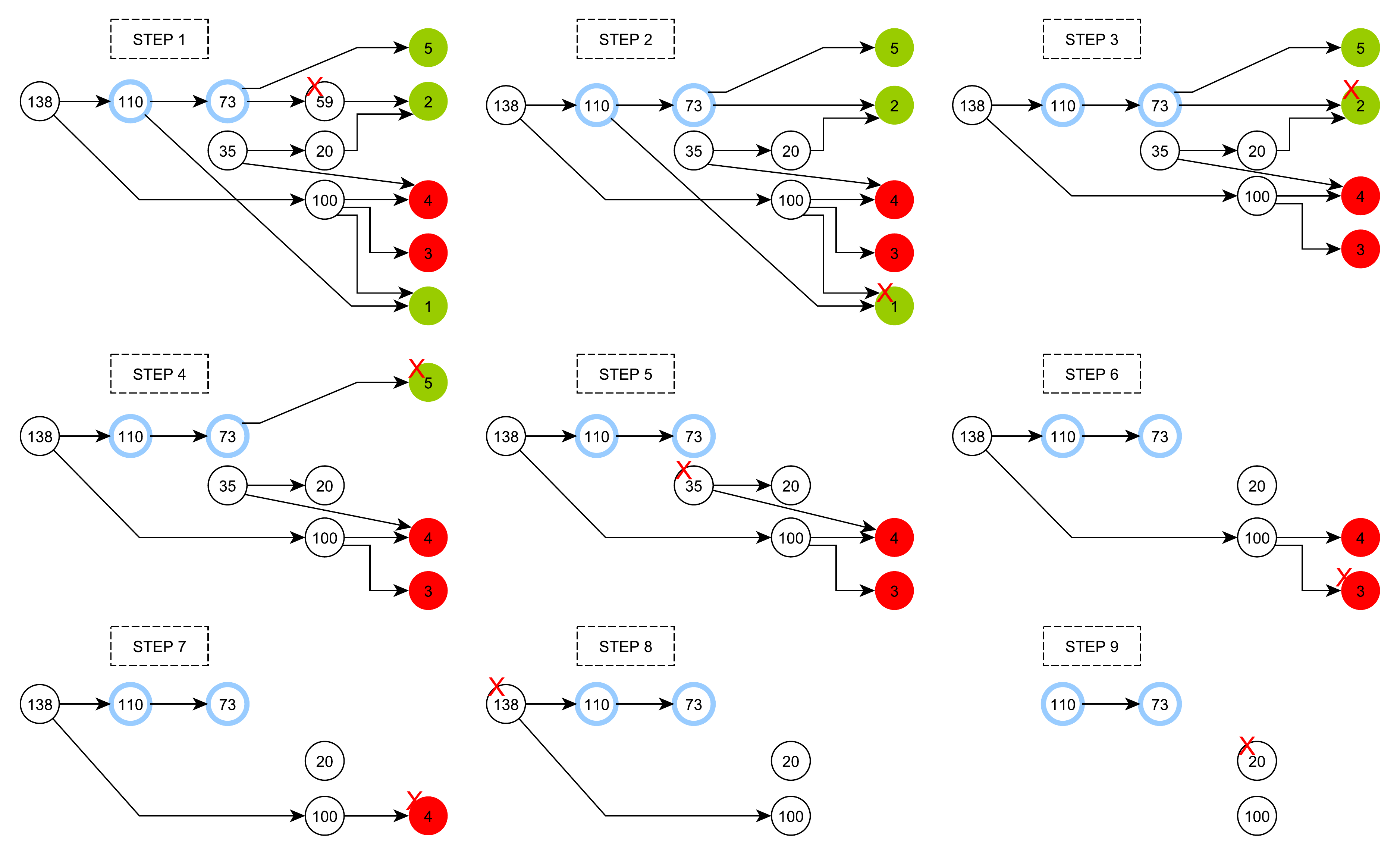} 
\end{center}
\caption{\emph{Coverage Graph} of the family problem. \darkgreen[Green] and \red[red] nodes refer to positive and negative examples respectively. Nodes with a \red[red] cross represent the candidate rules to be forgotten. Nodes with a thick \blue[blue] square represent those rules that have been consolidated. }
\label{fig:graphFamilySteps}
\end{figure*}

In order to clarify how this mechanism works we illustrate this with the \emph{Family} example. 
 Figure \ref{fig:graphFamilySteps} shows the evolution of the \emph{coverage graph} in Figure \ref{fig:graphFamily} and its measures (see Table \ref{fig:graphFamilySteps}) through nine consecutive forgetting steps, where the rule with lowest permanence is forgotten in each step (shown with a grey square). For instance, in step $1$, we see that rule number $59$ is redundant because it is covered by a more significant rule (with ID $110$), and it has the lowest value of permanence (see Table \ref{tab:tablefamilySteps} (step 1)). Thus, rule $59$ is forgotten, the  \emph{coverage graph} is redrawn (see  Figure \ref{fig:graphFamilySteps} (step 2)) and the metrics are recalculated if necessary (see Table \ref{tab:tablefamilySteps} (step 2)). 
In step $2$ (and other steps where a leaf node is deleted), its support is distributed equally among its ancestors and this distributed support becomes part of their \residual or intrinsic support ($\resc$). 

In this example, we have forgotten one rule at a time, but the actual pace and number of rules to forget can be tuned to the purpose of the system.

\begin{table*}[ht]
\begin{center}
\begin{tabular}{c}
\includegraphics[width=1.0\textwidth]{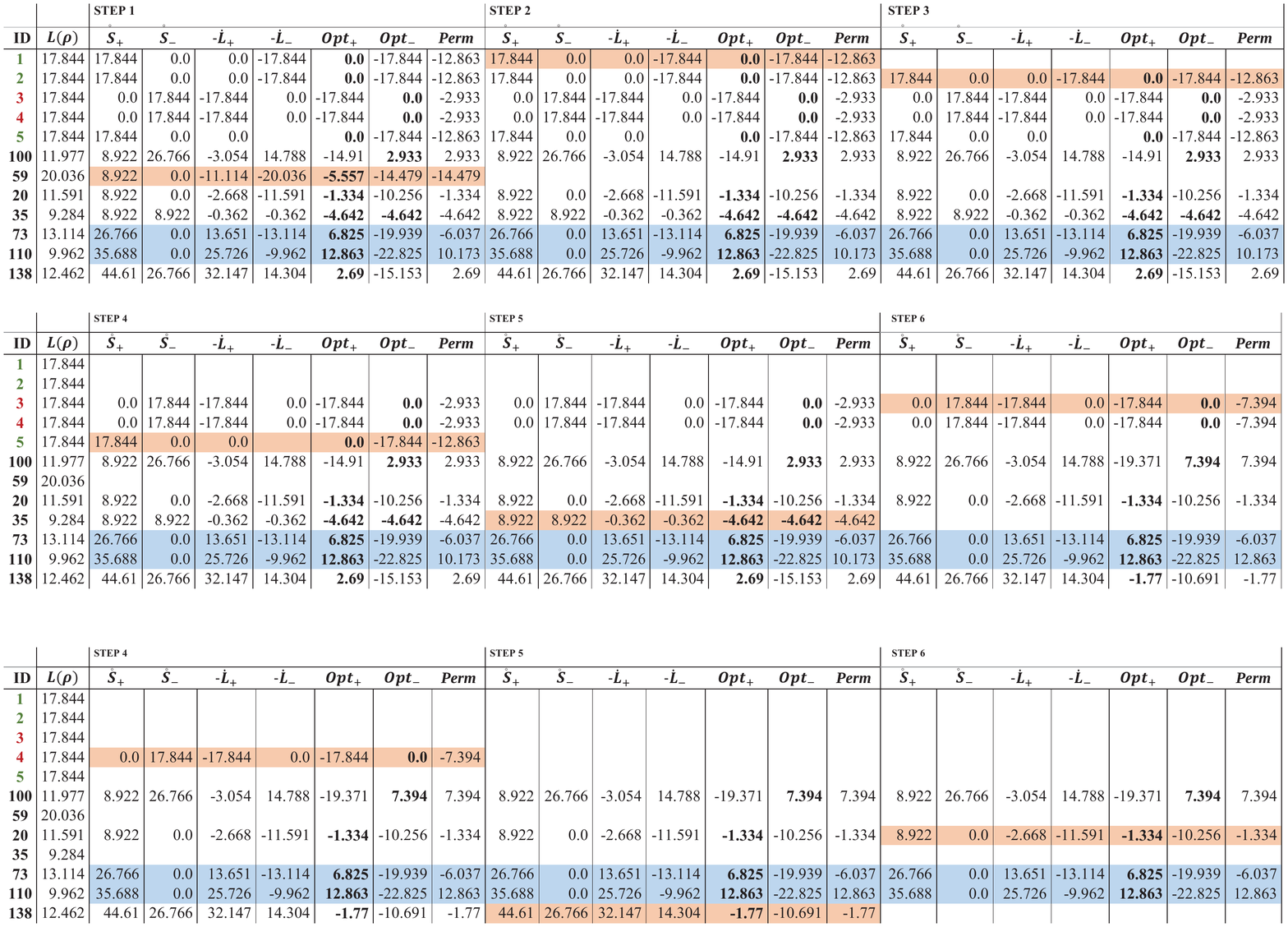}
\end{tabular}
\end{center}
\caption{Metrics for the family problem in nine steps of forgetting. Identifiers refer to the rules in Table \ref{tab:tablefamily}. Rows filled in \blue[blue] refers to those rules which have been consolidated. Rows filled in \orange[orange]  (non-consolidated rules with lowest $Perm$) refer to the rule candidate to be forgotten. }
\label{tab:tablefamilySteps}
\end{table*}%

\subsection{Consolidated knowledge: promotion and demotion} 

Finally, some of the rules with good indicators in the working space have to be eventually promoted to consolidated knowledge (or {\em belief}). This has to be a careful process, as the consolidated knowledge will be used by the deductive engine to calculate coverage. This means that an inconsistent rule that is promoted to the consolidated knowledge may have important consequences on the behaviour of the system.


The promotion function can be tuned for the application, but a general choice is to use a 
threshold $\theta_p$ on the \measureC to consolidate or promote a rule to a \emph{belief} status in $B$.

When a rule is promoted to consolidated knowledge, it cannot be target of the forgetting mechanism and, hence, be forgotten. 
It may happen that this rule can be eventually removed from the consolidated knowledge. Therefore, the promotion system is mirrored by a demotion system, with the use of another threshold $\theta_d$.
The original background knowledge ($B_0$) cannot be demoted (and forgotten).

In the example in Figure \ref{fig:graphFamilySteps}, we have established 
$\theta_p$ equal to the average \measureC of all the rules in the working space. Then, in step $1$, all the rules that exceed this average value will be consolidated to the background base (rules $110$ and $73$). Any rule that is 
consolidated cannot be target of the forgetting mechanism until it is demoted to the working space again (in the example, we have considered a demoting threshold $\theta_d$ equal to $\theta_p$). Thus, in Table \ref{tab:tablefamilySteps} (step $5$), rule $73$ has the lowest permanence value ($perm(73)=-6.037$) but $35$ ($perm(35)=-4.642$) is forgotten instead, because the former is a consolidated rule.

\section{Experiments}\label{sec:experimental}

As mentioned in section \ref{sec:intro}, one of the issues in many cognitive systems (especially connexionistic, either artificial or biological) is the Stability-Plasticity dilemma. We claim that our approach 
is able to address this issue in a long-life learning process. For this purpose, we have conducted an experimental evaluation to explore the following questions: (a) is it possible to gradually generate a large repository of consolidated knowledge assessing the usefulness of the rules? (b) is our approach able to forget or revise the existing knowledge in order to generate a rich and reusable knowledge base? and (c) how are the process and the resulting knowledge structure understood in terms of cognitive systems that must discover and develop knowledge incrementally?  We want to illustrate these features in one single domain. The ultimate goal of these experiments is to see whether the framework is general enough to work with off-the-shelf inductive and deductive engines, to better understand how the metrics and procedures work, and finding whether they may require some tuning or improvement to the framework before addressing other problems.

\subsection{Methodology}

\begin{figure}[htbp]
	\centering
		\includegraphics[width=0.3\columnwidth]{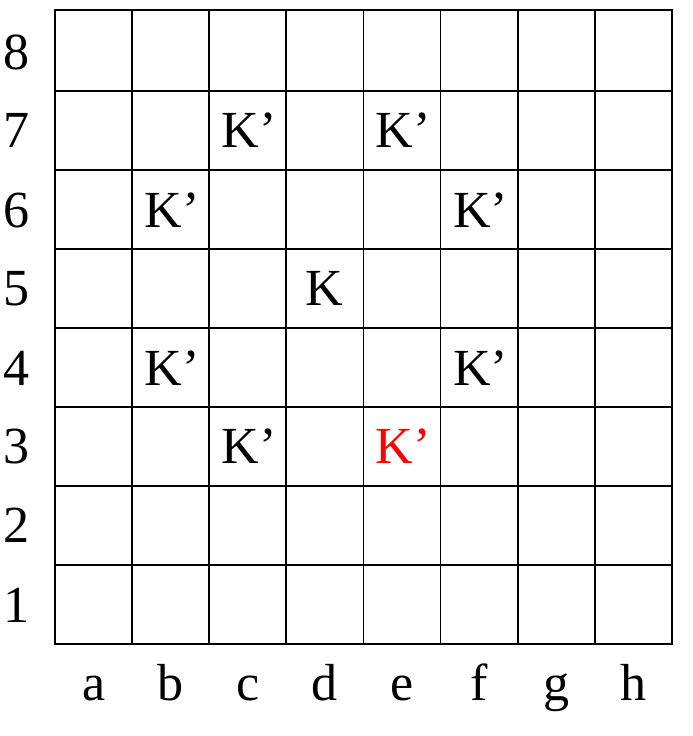}
		\caption{Possible moves of the knight from position (d,5). The particular legal move from $K$ to \red[$K'$] will be represented as {\ttfamily{move(knight,pos(d,5),pos(e,3))}}.}
	\label{fig:chessEx}
\end{figure}

We will focus on the problem of learning the rules of chess by observation. In particular, we focus on learning a model of legal moves of different pieces from a set of legal and illegal move examples (extracted from \cite{Muggleton89anexperimental}). 
In our framework, the legal moves are the positive examples and the illegal moves the negative ones (so we have two classes). Each example represents a move of a specific piece on an empty board. 
Therefore, a move is represented  by a triple from the domain $ Piece \times Pos \times  Pos $, where the second and third components represent, respectively, the piece's initial position and its destination on a chessboard. Positions are represented by a tuple from the domain $ File \times  Rank $ where files (a-h) stand for columns and ranks (1-8) stand for rows. For instance, Figure \ref{fig:chessEx} illustrates all the possible moves of a knight from a specific initial position ($K$) to several other positions ($K'$). We will use a {\sf Prolog} notation (as in the example in the previous section).
 
The only background predicate used is the absolute difference, {\em diff(X,Y)}, that calculates the distance between $X$ and $Y$, where both $X$ and $Y$ can be ranks or files (see Table \ref{tab:chessBK}).

\begin{table}[htbp]
\begin{center}
\begin{tabular}{c}
\includegraphics[width=0.6\columnwidth]{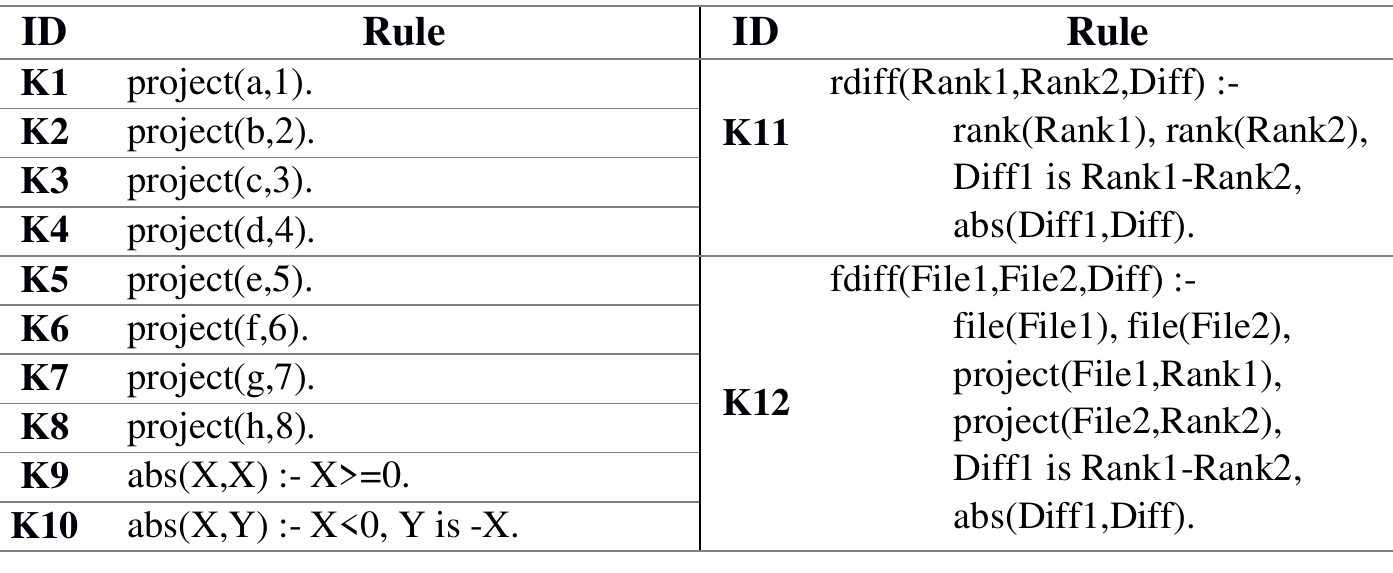}
\end{tabular}
\end{center}
\caption{Background knowledge for the chess problem.}
\label{tab:chessBK}
\end{table}%


The challenge we would like to face is knowledge discovery and acquisition in a progressive way from examples provided incrementally. A random set of chess moves from all chess pieces in the game except the pawn (rook, bishop, knight, queen and king) is given. This includes positive and negative examples (28 and 12 examples respectively). We also consider that an inductive engine is generating rules during the whole process (according to the  working space and using the consolidated knowledge as background knowledge) and they are arriving to the system in a random order as well. In our case, we have taken the rules generated by the ILP system {\sf Progol} \cite{progol} (60 in total). 
How many examples and rules are given for each step of the system is defined following a geometric distribution. Formally, the probability that $k$ examples (and similarly for rules) are given is $Pr(X=k)=(1-p)^{k-1}\cdot p$ where $k$ is $1, 2, 3, \dots$ and $p$ is the probability of success (we set it to $0.5$). In order to better mimic a situation where the inductive engine can produce rules it has already generated (as otherwise we would need to keep trace of all this), it is more realistic to use this distribution with replacement. Similarly, as the same move can appear repeatedly in chess, we have also considered replacement for the set of examples.


In this experiment, we have set the consolidation criterion with a threshold of \measureC greater than the average of the \measureC value of the rules in $W$ (provided that it is above the average \measureC of the evidence), 
namely, 

\begin{equation}\label{permChess}
\measureCab(\rho,W) > max(0,\frac{\sum\limits_{\nu \in W} \measureCab(\nu,W)}{|W|})
\end{equation}

\noindent Furthermore, since we want the consolidated knowledge to represent legal chess moves, we have set the $\beta$ parameter equal to $0.1$ in equation \ref{opt1} with the aim of penalising those rules that are not pure. 

\subsection{Consolidation without forgetting}

\begin{figure*}[!htbp]
\begin{center}
\includegraphics[width=1.0\textwidth]{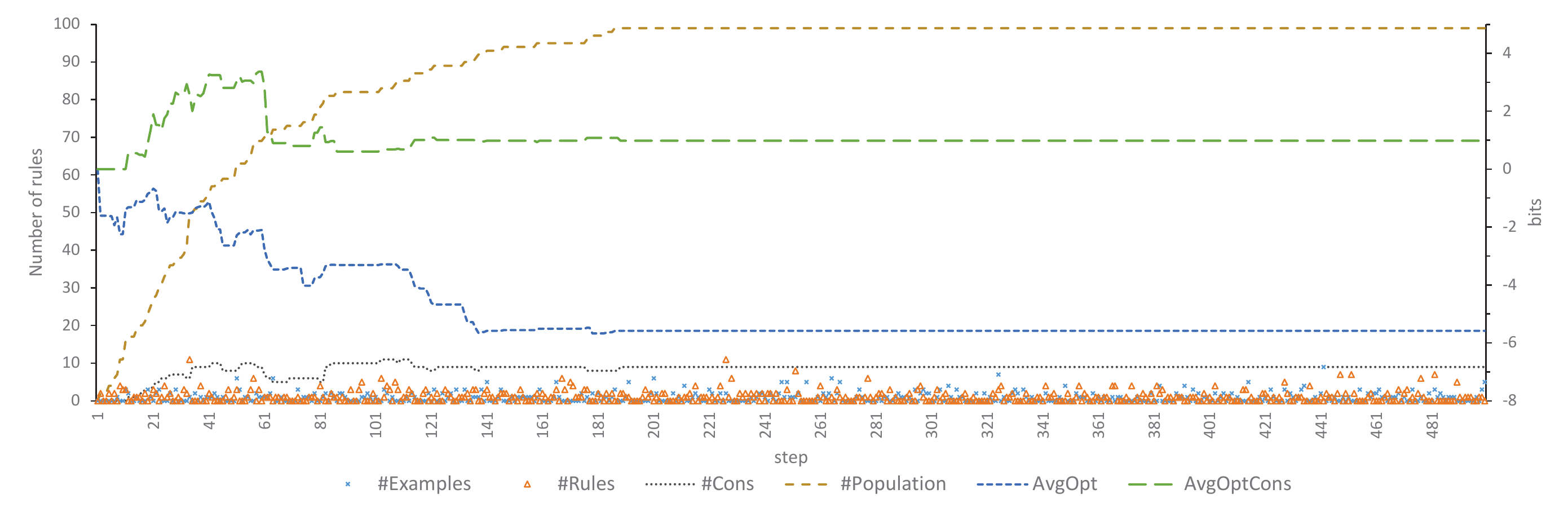} 
\end{center}
\caption{Evolution of some indicators for the chess problem \emph{without} the forgetting mechanism: {\em \blue[\#Examples]} and {\em \orange[\#Rules]} show the examples that arrive and the rules that are generated by the inductive engine for each step, \#Cons shows how many rules there are in the consolidated knowledge (initially the background knowledge) and {\em \brown[\#Population]} shows the total number of rules (magnitudes shown on the left \yaxis). {\em \darkblue[AvgOpt]} and {\em \darkgreen[AvgOptCons]} show, respectively, the average optimality for all rules and  the average optimality for all consolidated rules (magnitudes shown on the right \yaxis). This Figure shows how, after the working space is filled with all the evidence and generalised rules, the metrics become stable.
}
\label{fig:graphA}
\end{figure*}

\begin{figure*}[!htbp]
\begin{center}
\includegraphics[width=1.0\textwidth]{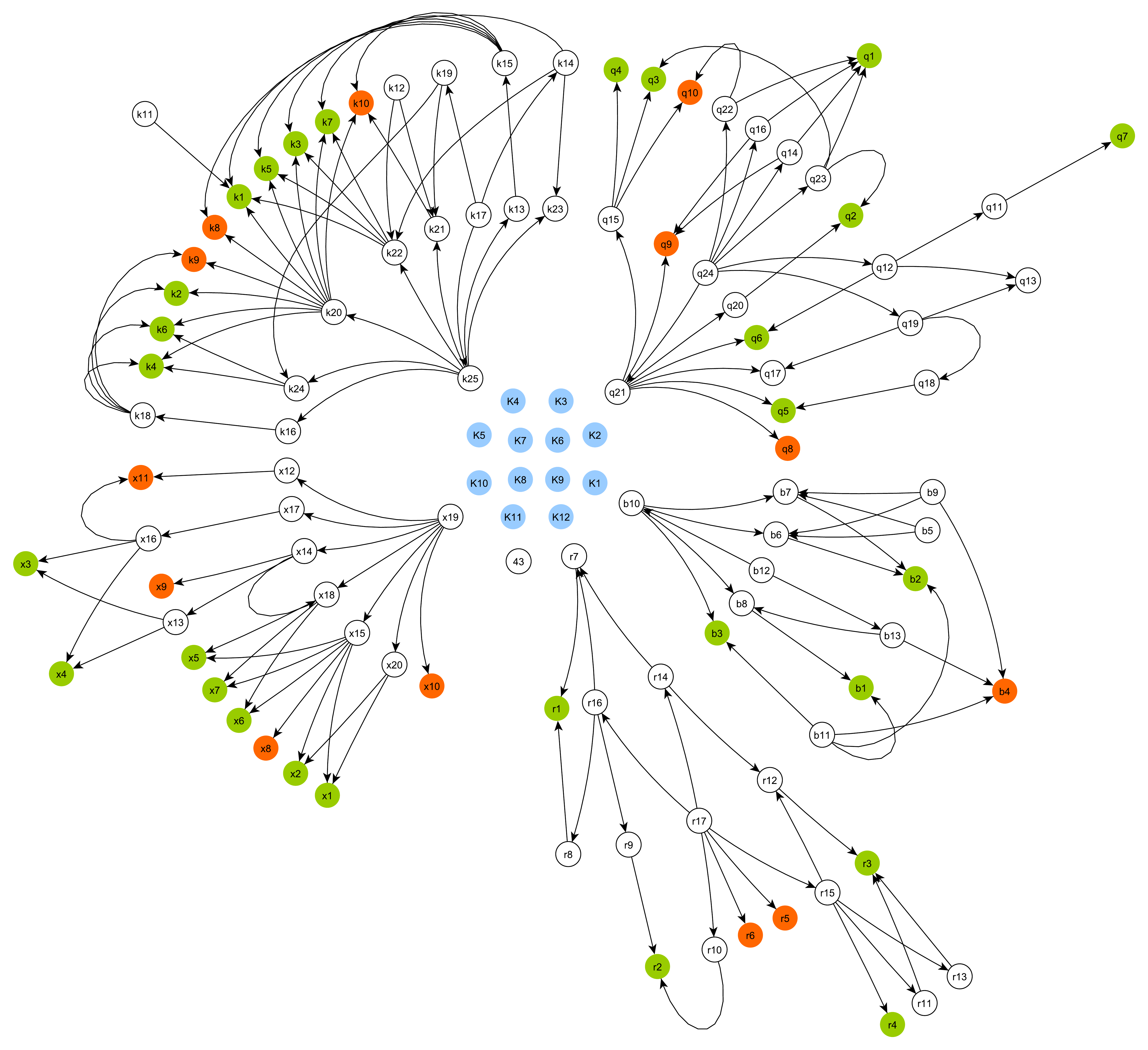} 
\end{center}
\caption{\emph{Coverage graph} that represents the coverage relations between the individuals for the chess problem (without oblivion) at step $400$. The metrics are shown in Table \ref{tab:tableAall}. \darkgreen[Green] and \red[red] nodes refer to positive and negative examples respectively. Original background knowledge is represented as blue nodes.}
\label{fig:graphAall}
\end{figure*}

In a first experiment we try to show what would happen without applying the forgetting mechanism and check whether the MML-based measures 
work successfully for knowledge acquisition: are the final consolidated knowledge useful to solve the problem given the evidence? Figure \ref{fig:graphA} shows the evolution of the learning process during 500 steps. As no rules are forgotten, the rule population (dashed \brown[brown] line) reaches its maximum value (100) and it stagnates ignoring any new evidence which arrives to the system (because they are already placed in $W$) from step 180 onwards. In this case we have assumed that all the evidence of the chess problem can be allocated in $W$, however it could be the case that all knowledge of a problem will not fit into $W$ (memory restrictions) thus collapsing with no improvement. 
 The same applies to both the average optimality of all rules (dashed \darkblue[blue] line) and the consolidated ones (dashed \darkgreen[green] line) which, since no more new rules are allocated into $W$, no further learning or knowledge improvement can take place. Table \ref{tab:tableA} shows the consolidated rules at step $500$ where we can see that they almost represent all the legal chess moves (only two movements of the knight are missing in this set) and there is only one rule ($x20$) which, despite representing a legal move, does not completely generalise the movement of the piece (king). 
This is a good result as the working space is large enough to accommodate all these rules (and many other less significant rules).
See Table \ref{tab:tableAall} in Appendix \ref{ApRules} for all the rules in $W$ at step $500$ and Figure \ref{fig:graphAall} for their coverage relations. The conclusion we can draw from these results is that the metrics used to measure the usefulness of the rules provide a guarantee of promoting those rules that, having the maximum compression, best describe the problem.



\begin{table}[!htbp]
\begin{center}
\begin{tabular}{c}
\includegraphics[width=0.9\textwidth]{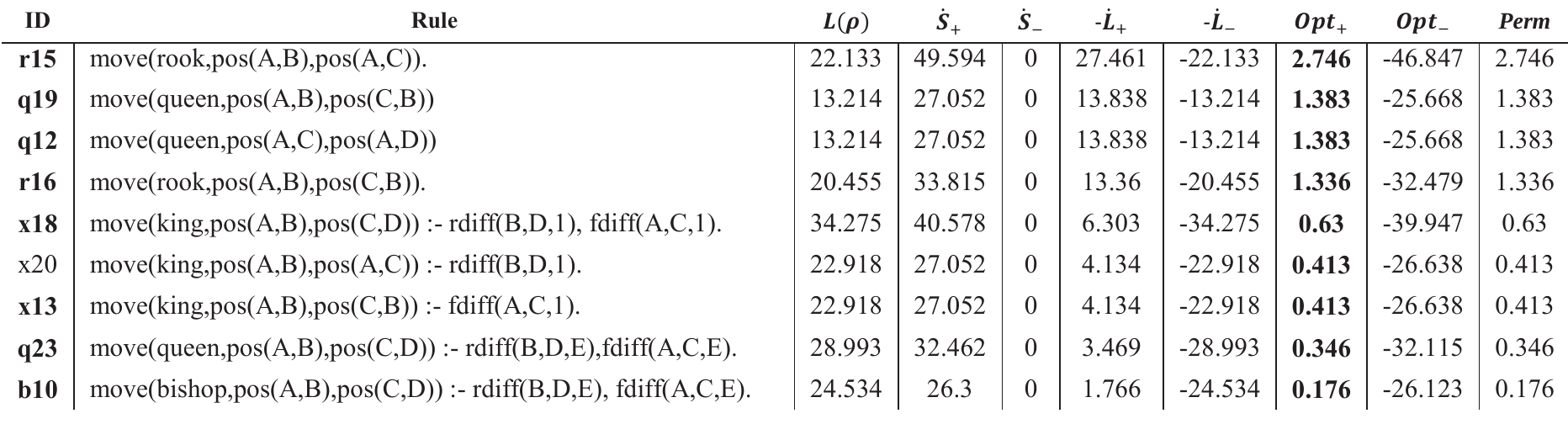}
\end{tabular}
\end{center}
\caption{Consolidated rules and metrics for the chess problem \emph{without} the forgetting mechanism at step $500$. IDs in bold represent those rules that perfectly generalise the legal moves of the chess pieces. 
}
\label{tab:tableA}
\end{table}%

\subsection{Consolidation with forgetting}

After that, we repeat the same experiment, but using the forgetting mechanism. This tries to represent a situation where we have bounded resources, in this case a more limited working space, so it is necessary to forget rules in order to allocate new ones. 
What we want to show is that if our approach is able to find a solution to a certain problem without the use of the forgetting mechanism, a suitable (and possibly better) solution to the problem should exist having bounded resources and by using the forgetting mechanism. 
In order to do that, we have executed several configurations with varying maximum number of rules in the working space ($|W| \in \{(20, 30, 40, 50, 60, 70, 80, 90)\}$) and every time the limit is exceeded the forgetting process is launched, forgetting up to 25\%, 50\% or 75\% of the most meaningless rules (those with the lowest $perm$ value). Each different configuration has been launched 10 times, 
hence, there are $240$ executions in total.

\begin{table*}[!htbp]
\begin{center}
\begin{tabular}{c}
\includegraphics[width=0.8\textwidth]{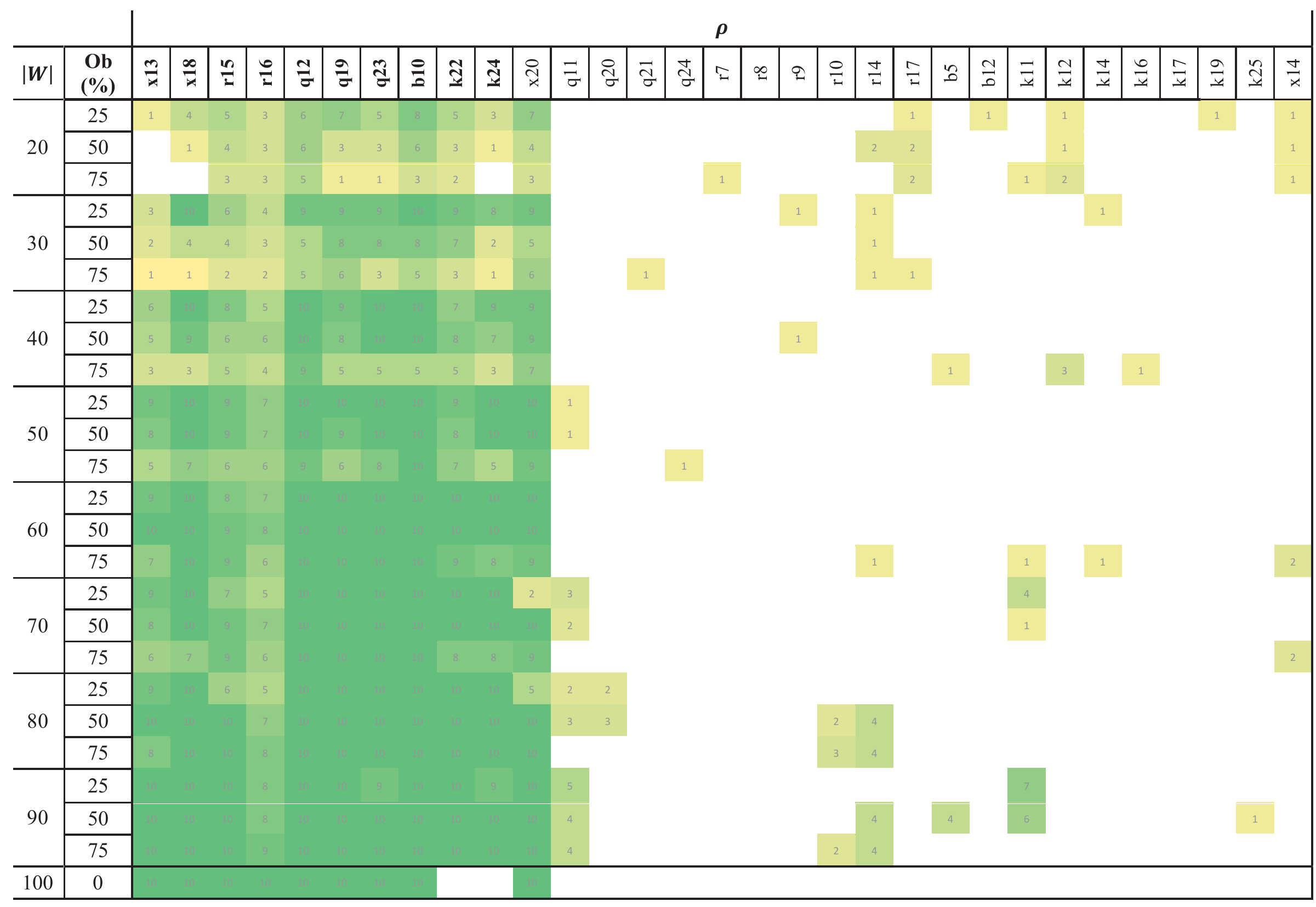}
\end{tabular}
\end{center}
\caption{Heat map showing the percentage of times a rule has been consolidated for each different configuration (maximum number of rules, 20-100,   percentage of rules forgotten, 25\%-75\%). The last row shows the results without forgetting. 
Each cell represents 10 repetitions. 
The latter row ($|W| = 100$) represents the reference solution, namely, the solution obtained by the experiment without forgetting (previous section). Rules ($\rho$) in bold are those rules that belong to the solution of the problem. As it can be seen, even with very limited resources, the consolidated knowledge improves the reference solution. 
}
\label{tab:tableExp}
\end{table*}%

Table \ref{tab:tableExp} is a \emph{Heat map} showing, for each possible configuration ($|W| \times forgetting (\%)$) how many times a specific rule appears in the consolidated knowledge in 10 repetitions, from white ($0$ times), 
light yellow ($1$ time) to dark green ($10$ times). Rules that are not represented in the Heat Map is because they have not been consolidated at any time.
Knowing that the consolidated rules by the first experiment (Table \ref{tab:tableA}) are those represented in the bottom row ($|W| = 100$), it is easy to see that not only the set of consolidated rules almost always includes the reference solution (even with very limited resources), but also the forgetting criterion allows the system to include those rules that perfectly generalise the moves of the king (rules in bold). The rest of rules included in the consolidated set in each experiment also generalise different movements of the pieces and, in some cases, they could disappear from this set by using a more restrictive consolidation criterion (i.e., by using the average of the \measureC plus $n$ times its standard deviation). See Table \ref{tab:tableBall} in Appendix \ref{ApRules} for all the rules in $W$ at step $500$.

%

\begin{figure*}[!htbp]
\begin{center}
\includegraphics[width=1.0\textwidth]{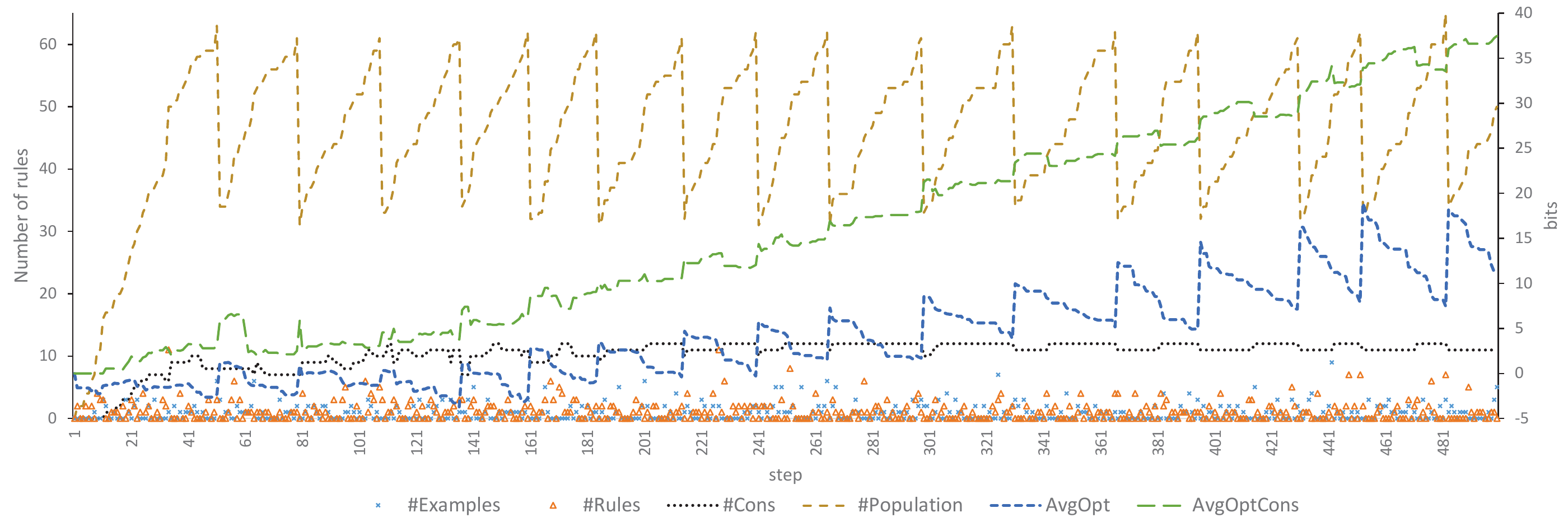} 
\end{center}
\caption{Evolution of the same indicators as in Figure \ref{fig:graphA} for the chess problem {\em with} the forgetting mechanism (for a configuration  with maximum number of rules 60 and up to 50\% of rules forgotten for each forgetting step). Now we see a bumpier picture, where the forgetting mechanism takes place every 30 steps approximately.}
\label{fig:graphB}
\end{figure*}

In order to compare both experiments, Figure \ref{fig:graphB} shows the evolution of the system during 500 steps for one representative setting of the 
24 configurations (maximum number of rules equals to 60 and up to 50\% of rules forgotten in each forgetting step). Now, the variations in the amount of consolidated rules  (dotted black line) and rules in the working space (dashed \brown[brown] line) allow us to observe how the forgetting mechanism works (every 30 steps approximately).
Table \ref{tab:tableB} presents the consolidated rules at the final step ($500$). In this case, this set perfectly generalises all the legal moves of all the chess pieces. The system has reached a stable situation in which the number of consolidated rules (dotted black line) remains almost constant from step $250$. 
The average optimality of both the consolidated rules (dashed \darkgreen[green] line) and all the rules (dashed \darkblue[blue] line) have an increasing trend due to the distribution with replacement used to populate the working space. The appearance of new rules in the system or the execution of the forgetting mechanism mainly affect the average optimality of $W$ (dashed \darkblue[blue] line): every time it runs, the working space is cleaned of useless rules which strongly affects the metrics of the rules in $W$ (and to a lesser extent to the consolidated set of rules (\darkgreen[green] line)) that have to be recalculated. 
Compared with the former experiment, the number of rules in $W$ has been reduced (with one order of magnitude (10x) speedup in execution) obtaining a better set of consolidated knowledge: it includes all the rules that solve the chess problem, including the two legal moves of the knight, rules $k22$ and $k24$, which were missing from the consolidated knowledge in the first experiment.

\begin{table*}[!htbp]
\begin{center}
\begin{tabular}{c}
\includegraphics[width=0.9\textwidth]{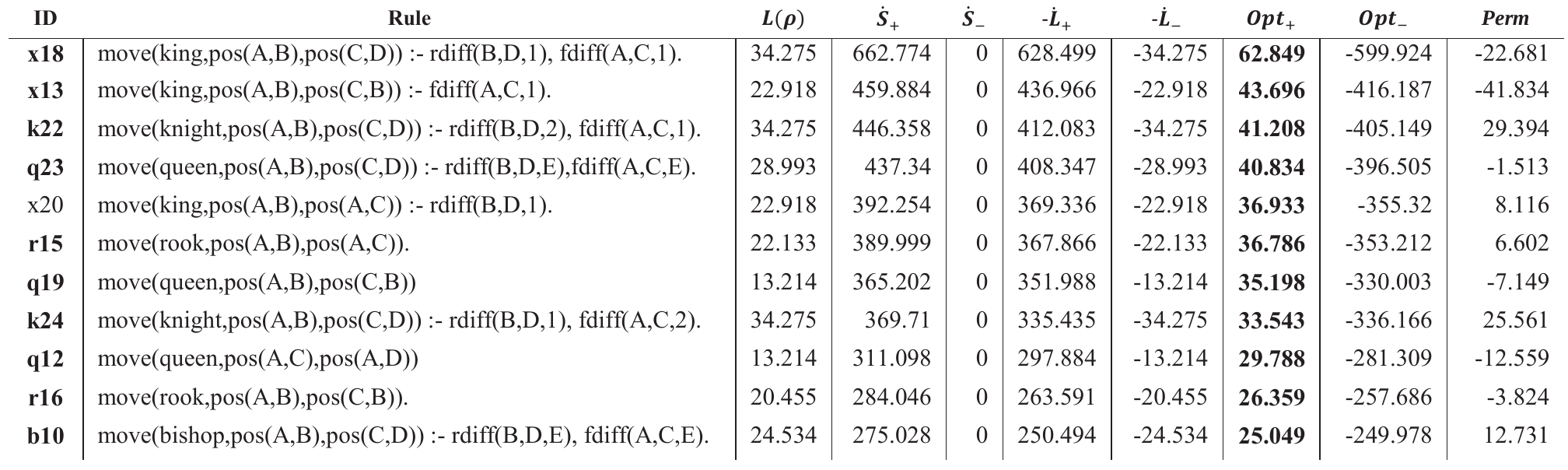}
\end{tabular}
\end{center}
\caption{Consolidated rules and metrics (as in Table \ref{tab:tableA}) for the chess problem \emph{with} the forgetting mechanism at step $500$ (for a configuration  with maximum number of rules 60 and up to 50\% of rules forgotten for each forgetting step). IDs in bold represent those rules that perfectly generalise the legal moves of the chess pieces. 
}
\label{tab:tableB}
\end{table*}%

\subsection{Incremental knowledge acquisition}

\begin{figure*}[!htbp]
\begin{center}
\includegraphics[width=1.0\textwidth]{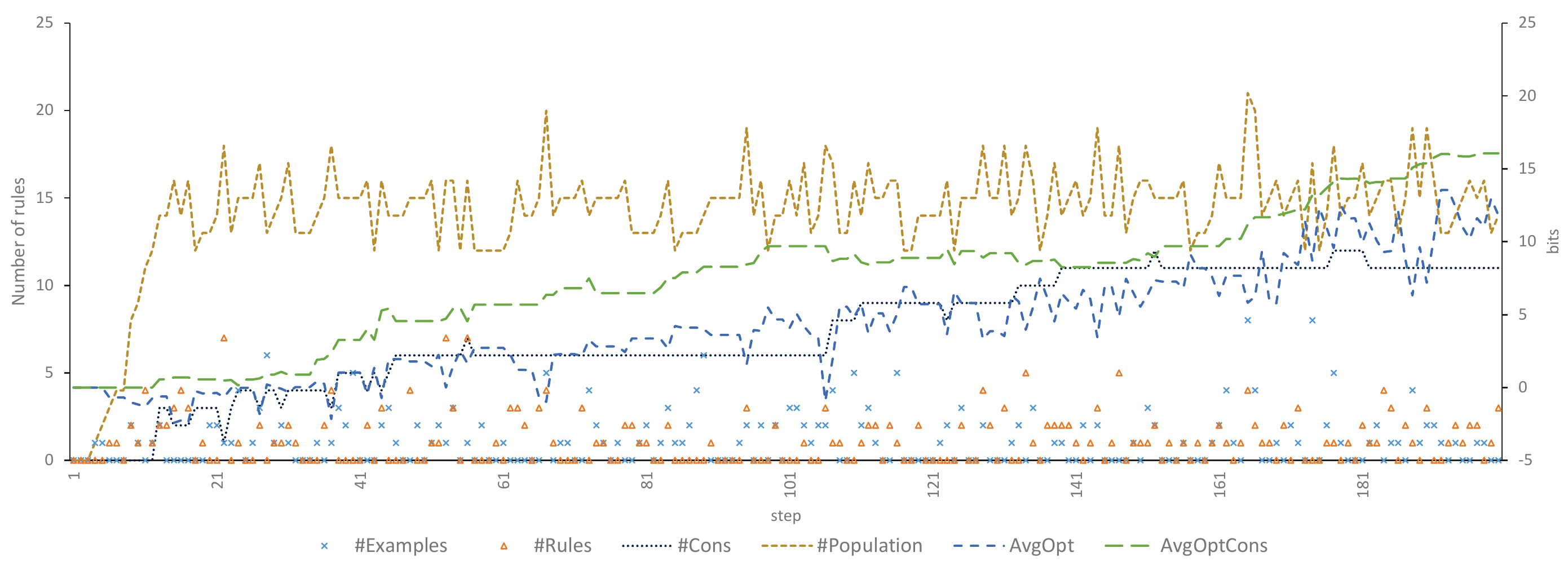} 
\end{center}
\caption{Evolution of the same indicators as in Figure \ref{fig:graphA} for the incremental chess problem (rook and bishop moves in the first $100$ steps, and queen moves in the following $100$ steps) {\em with} the forgetting mechanism. We see a non-constant sawtooth-like picture for the number of rules in the working space where the forgetting mechanism takes place every little number of steps due to the small amount of rules allowed and the low percentage of rules forgotten in every forgetting step. Nonetheless, the consolidated rules became constant in each different learning process.}
\label{fig:graphReuse}
\end{figure*}

Finally, one last experiment tries to show the capability of our approach for the incremental learning of new knowledge from previously consolidated concepts. 
This experiment is divided in two phases: in the first one we have only taken rules and examples of moves of the rook and bishop chess pieces (15 and 30 rules respectively) providing the system with them in the same way as in the previous experiment. The consolidation criterion has not been changed, but the maximum number of rules in the working space has been established to 15 (in order to allow the forgetting mechanism to work) and the percentage of meaningless rules that are forgotten for each forgetting process up to $25\%$, due to the smaller size of the working set. In Table \ref{tab:tableReuse100} we can see the set of consolidated rules after $100$ steps. This set contains the rules that perfectly generalise all the legal moves of the rook and the bishop. In the first 100 steps of Figure \ref{fig:graphReuse} we can see how the forgetting and consolidation mechanisms work. This time, due to the lower maximum number of rules allowed in the working space, the lower percentage of rules forgotten and the geometric distribution used to provide the rules, the forgetting mechanism runs here every few steps, showing non-constant sawtooth-like wave ramps for the number of rules in the working system (dashed \brown[brown]). However, the number of consolidated rules remains constant almost from step 45 to the end of this stage (100).

\begin{table*}[!htbp]
\begin{center}
\begin{tabular}{c}
\includegraphics[width=0.9\textwidth]{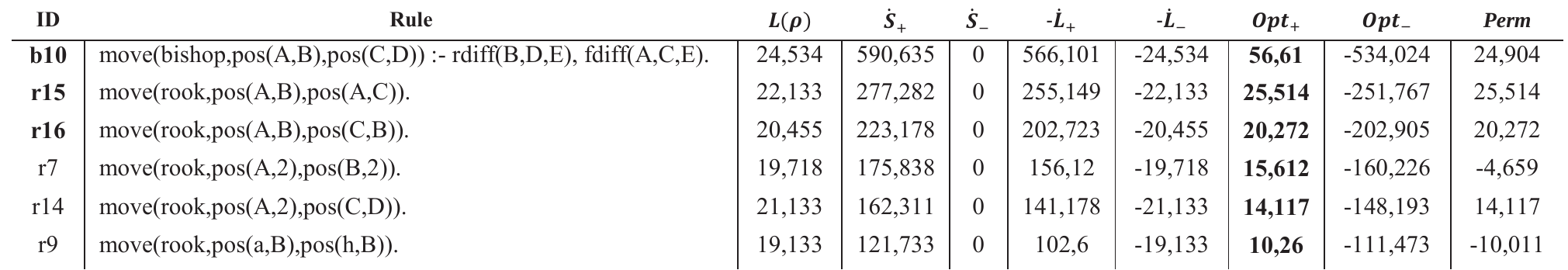}
\end{tabular}
\end{center}
\caption{Consolidated rules and metrics (as in Table \ref{tab:tableA}) for the chess problem (rook + bishop moves) at step $100$. All rook and bishop legal moves are covered by these rules and no better rules can be obtained.}
\label{tab:tableReuse100}
\end{table*}%

In the second phase, we provided the system with a new set of rules and examples (10 and 20 rules respectively) only representing moves of the queen chess piece. Apart from using the background knowledge that is provided initially, it should also be possible at this point to use the previously learned moves of the rook and the bishop in order to express the moves of the queen. This is what the inductive engine can take advantage of. Table \ref{tab:tableReuse200} shows the set of consolidated rules which contains the previously consolidated rules that generalise the legal moves of the rook and bishop, and a new set of rules that represents the legal moves of the queen. This latter set includes a pair of rules ($q29$ and $q25$) that use the rook and bishop rules and represent all the possible moves of the queen piece: $q25$ which covers both the horizontal and vertical moves of the queen; and  $q29$ which covers the diagonal movement. The second half of Figure \ref{fig:graphReuse} (from step 100) shows how the forgetting mechanism runs even more frequently than previously (dashed \brown[brown]) due to the increment of consolidated rules (that cannot be targeted by forgetting). Again, the number of consolidated rules (dotted black line) remains constant most of the time (from step 140 to step 200).

\begin{table*}[!htbp]
\begin{center}
\begin{tabular}{c}
\includegraphics[width=0.9\textwidth]{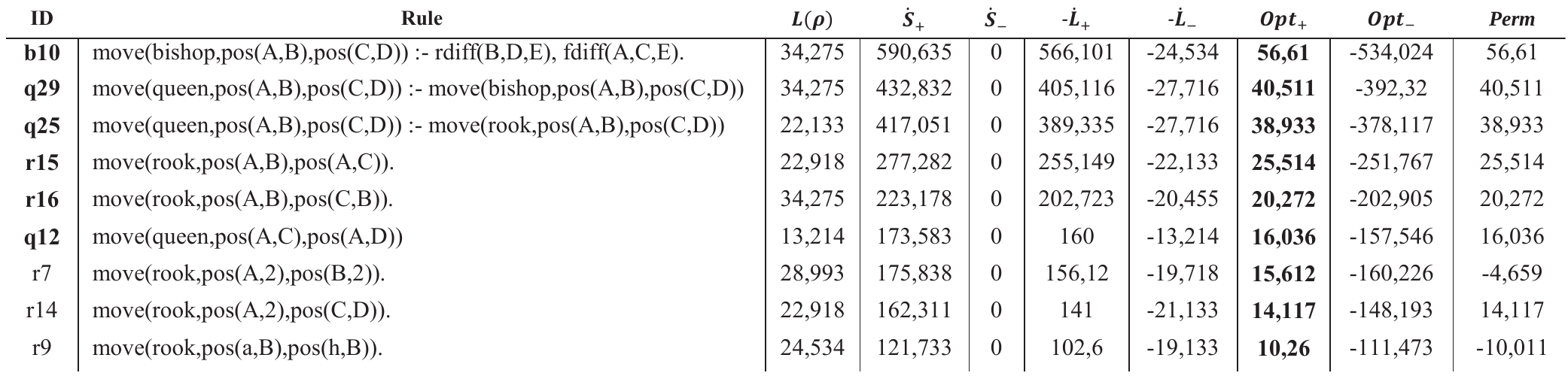}
\end{tabular}
\end{center}
\caption{Consolidated rules and metrics (as in Table \ref{tab:tableA}) for the chess problem (queen moves) at step $200$ (the $100$ firsts steps for learning the rook and bishop moves, and the $100$ following steps for learning the moves of the queen). All legal moves of the queen are covered by taking advantage of previously learned moves of the rook and bishop, whose legal moves are also covered by this set.}
\label{tab:tableReuse200}
\end{table*}%

\subsection{Discussion}

In an effort to facilitate an understanding of whether our approach is able to effectively and incrementally grow a knowledge base by using appropriate evaluation metrics and useful cognitive abilities for addressing the knowledge acquired, we have performed some experiments over a well-known scientific domain, the chess problem. As we have said, the ultimate goal is not to validate the approach but to provide some insight into both its generality, efficiency and the much-needed use of forgetting and consolidation cognitive procedures in incremental and developmental approaches for knowledge discovery. In order to shed some light on these aspects, we will refer to the questions raised at the beginning of this section.

From the above experiments, we see that the repository of rules can be well structured and ranked by the metrics and the system consolidates those rules that are appropriate, therefore responding affirmatively to the first question (a). Regarding question (b), we also see that a moderate limitation of working space with forgetting is even capable to improve the identification of the rules to be consolidated, and, what is better, prevents the system for stagnating or collapsing in situations where we have bounded resources. Finally, in connection with question (c), we see the behaviour in an incremental setting, where the knowledge can be used in new tasks; one of the principles of developmental cognition.

Consequently, the proposed approach for knowledge acquisition is a favourable compromise to the stability-plasticity dilemma, which is characterised as:

\begin{itemize}

\item Too much plasticity will result in previously learned knowledge being constantly forgotten. However, the promotion and demotion mechanisms together with the evaluation metrics rank and structure the knowledge allocated in the working space avoiding useful knowledge losses.  

\item Too much stability will impede the efficient coding of new learnt knowledge. However, the forgetting mechanism also together with the evaluation metrics is in charge of removing those meaningless and redundant knowledge.

\end{itemize}



\section{Conclusions}\label{sec:conclusions} 

Learning a set of rules from data is nowadays a well-known problem for which many approaches exist. However, the use of background knowledge and the consolidation of new knowledge is one of the conspicuous problems in the understanding and creation of cognitive systems, and the management of more long-life knowledge discovery systems. The organisation of complex knowledge structures in terms of coverage graphs allows a straightforward and principled approach to knowledge acquisition, consolidation (promotion), revision (demotion) and forgetting. All this can be applied and analysed at a meta-level, with the use of off-the-shelf deductive and inductive engines. This modularity, and the ability of dealing with declarative knowledge bases opens up a range of applications in knowledge discovery, developmental cognition, expert systems and other intelligent systems that are meant to have a non-ephimeral life.

The main contributions of this work are: (1) The first extension of the MML principle to a knowledge network (in the form of coverage graph). While the MML principle has a Bayesian inspiration, the metrics are more flexible than actual probabilities, stauncher when pieces of the working space are removed, and can be combined into metrics for different processes. (2) 
We show that the development of a formal epistemology to support knowledge discovery, in terms of how the knowledge can be acquired and justified, supports a constructive and developmental way to  define appropriate knowledge acquisition processes. In particular, we have seen how cognitive procedures as the forgetting criterion are not only necessary when the working space is finite but it can even be beneficial in our setting. (3) Our approach is parametrisable to other cognitive or intelligent systems, as it works at a meta-level and is independent of the actual deductive and inductive mechanisms that are used underneath. (4) The nonmonoticity problem of knowledge acquisition and revision is approached in a more lightweight and robust way, and the system can cope with redundancy and even inconsistency without heavy conflict resolutions or complex semantic artifacts. (5)  
The problem  of catastrophic forgetting and, thus, \emph{The Stability-Plasticity} dilemma has been effectively overcome when acquiring knowledge allowing to our approach not only gain new knowledge, but also addressing it efficiently. (6) Its adaptive an off-the-shelf characteristics allow to feed our approach on dynamic data in real time, or near real time.

Given the flexibility of the approach we consider many avenues of future work. We plan to apply the setting to some other applications, by using the same or other deductive and inductive engines, and keep on with the integration into our learning system \gerl \cite{gerlNFMCP12}. 
Furthermore, it is also of interest the application of the principles used (MML evaluations and cognitive mechanisms) in other kind of AI systems such as decision support systems in order to help them make better decisions based on the best available data. Finally, two further desirable characteristics for our approach are also likely to be part of our future research: (a) interactiveness, namely, the ability to find an additional (human or not) source input if a problem statement is ambiguous or incomplete; (b) contextuality, in terms identify, understand and extract contextual elements such as syntax, semantics, domain , time, location, goal, \dots, which may be useful to move beyond the current knowledge acquisition systems.


\section*{Acknowledgements}
This work has been partially supported by the EU (FEDER) and the Spanish MINECO under grants TIN 2010-21062-C02-02, TIN 2013-45732-C4-1-P and FPI-ME grant BES-2011-045099, and by Generalitat Valenciana PROMETEO2011/052.

\vspace{0.2cm}
\noindent Part of this work is under consideration at Pattern Recognition Letters.

\bibliographystyle{IEEEtran}
\bibliography{biblio}

\newpage
\section{Appendix}\label{ApRules}

\begin{table*}[!htbp]
\begin{center}
\begin{tabular}{c}
\includegraphics[width=0.66\textwidth]{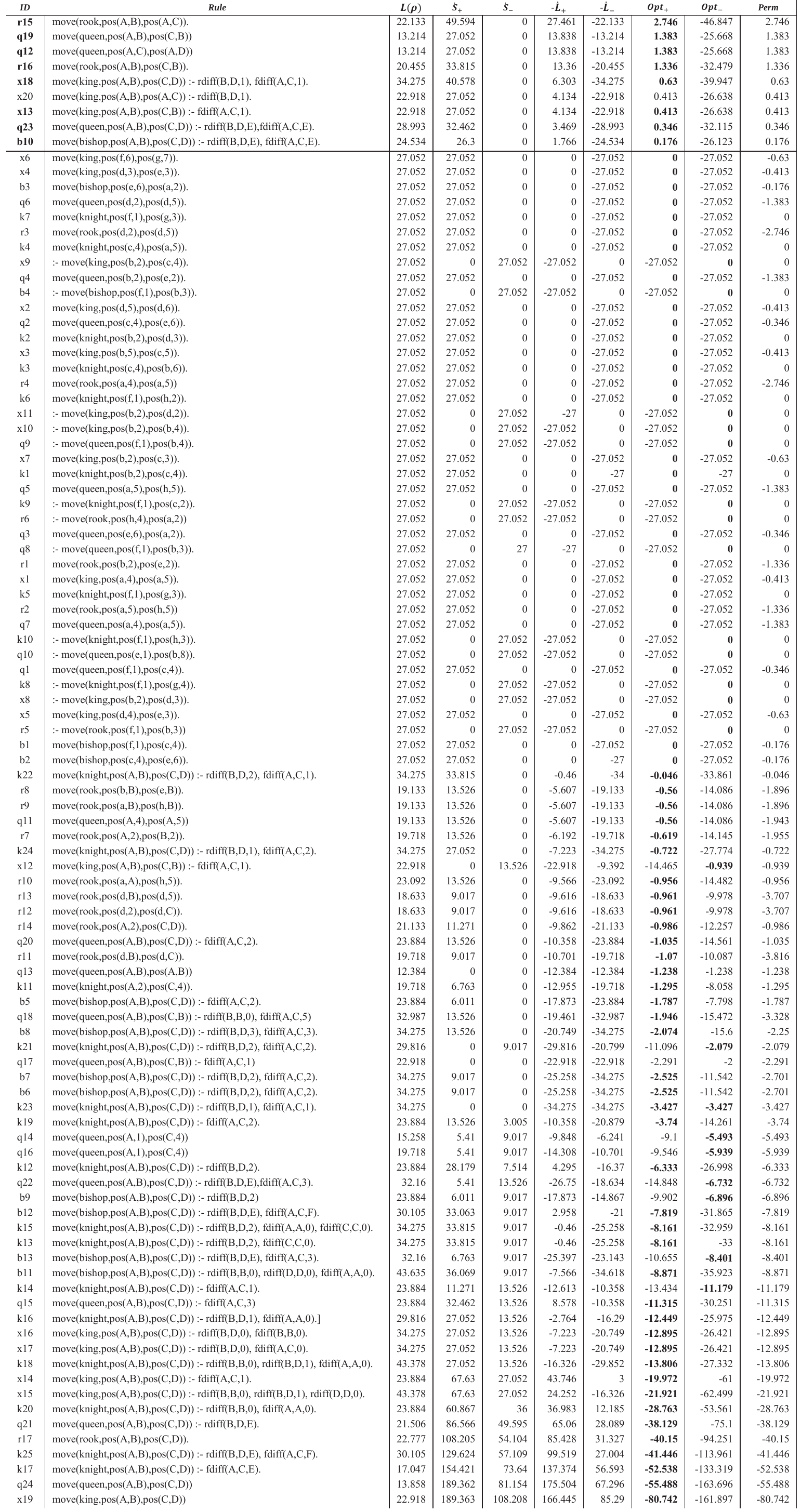}
\end{tabular}
\end{center}
\caption{The complete set of rules and their metrics (same as in Table \ref{tab:tableA}) for the chess problem \emph{without} the oblivion mechanism at step $500$. Consolidated rules are placed in the table at the top while the rest of rules in $W$ are placed in the table at the bottom.}
\label{tab:tableAall}
\end{table*}%

\begin{table*}[!htbp]
\begin{center}
\begin{tabular}{c}
\includegraphics[width=1.0\textwidth]{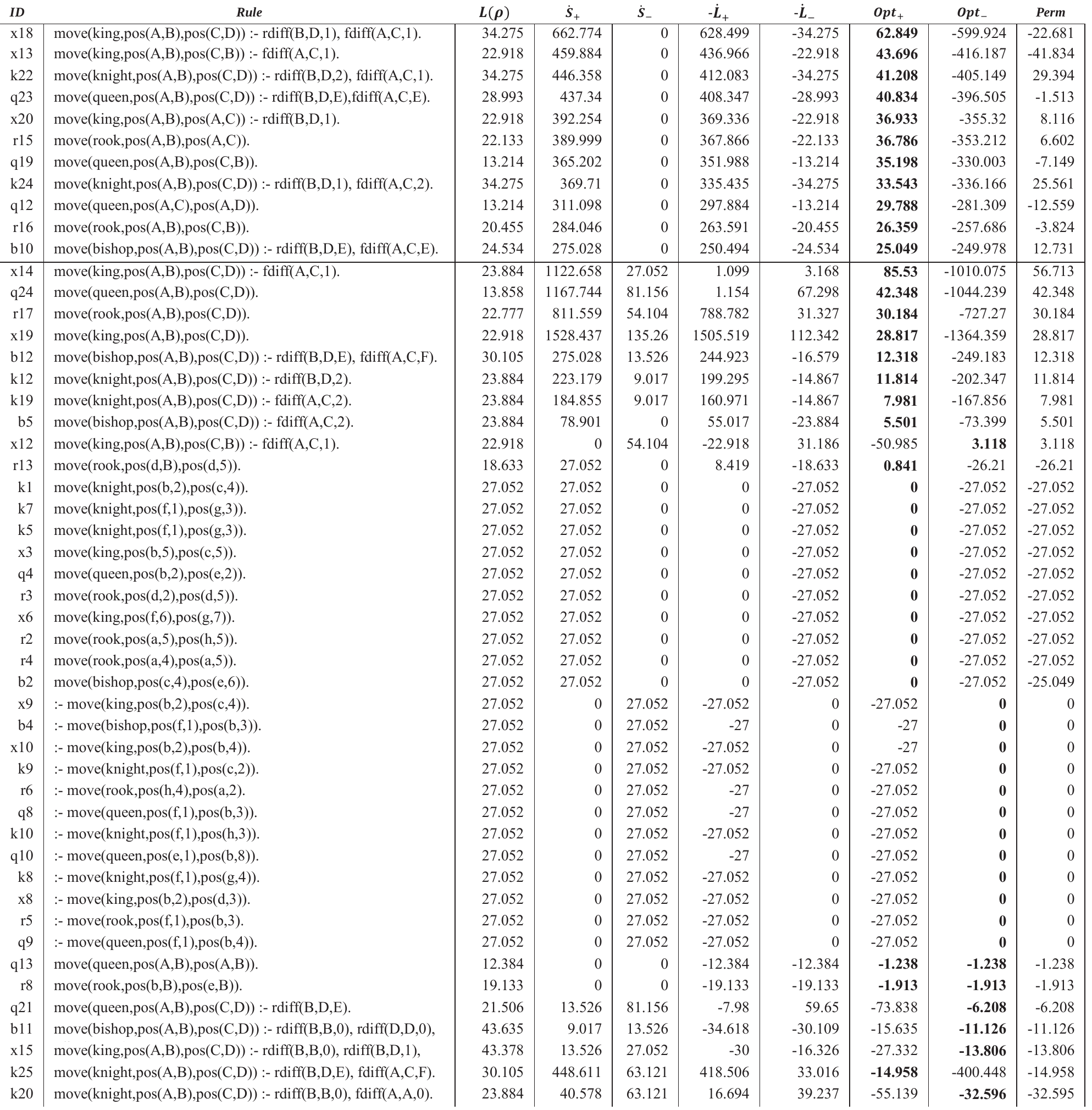}
\end{tabular}
\end{center}
\caption{Rules and metrics (same as in Table \ref{tab:tableA}) for the chess problem \emph{with} the oblivion mechanisms at step $500$.  Consolidated rules are placed in the table at the top while the rest of rules in $W$ are placed in the table at the bottom.}
\label{tab:tableBall}
\end{table*}%


%





\end{document}